\pdfoutput=1

\documentclass[11pt]{article}

\usepackage[]{acl}

\usepackage{times}
\usepackage{latexsym}
\usepackage{booktabs}
\usepackage{xspace}
\usepackage{longtable}
\usepackage{graphicx}
\graphicspath{ {./images/} }
\usepackage{csvsimple}
\usepackage{multirow}

\usepackage[T1]{fontenc}

\usepackage[utf8]{inputenc}

\usepackage{microtype}

\usepackage{inconsolata}

\usepackage{tikz}
\usetikzlibrary{shapes.geometric, arrows}
\usepackage{csquotes}
\usepackage{tablefootnote}
\usepackage{float}

\newcommand{\belebele}{\textsc{Belebele}\xspace}

\newcommand{\nlangs}{122\xspace}

\newcommand{\nfams}{27\xspace}
\newcommand{\falcon}{\textsc{Falcon}\xspace}
\newcommand{\llama}{\textsc{Llama}\xspace}
\newcommand{\llamatwochat}{\textsc{Llama-2-chat}\xspace}
\newcommand{\gptturbo}{\textsc{GPT3.5-turbo}\xspace}
\newcommand{\flores}{\textsc{FLoRes}\xspace}

\title{The \belebele Benchmark: \\
a Parallel Reading Comprehension Dataset in 122 Language Variants}

\author{Lucas Bandarkar\textsuperscript{*$\mathsection$}, \ Davis Liang\textsuperscript{*$\dagger$}, \ Benjamin Muller\textsuperscript{*}, \\
{\bf Mikel Artetxe}\textsuperscript{*$\ddagger$}, {\bf Satya Narayan Shukla}\textsuperscript{*}, {\bf \ Donald Husa}\textsuperscript{*}, {\bf Naman Goyal}\textsuperscript{*}, \\ 
 {\bf Abhinandan Krishnan}\textsuperscript{*}, {\bf Luke Zettlemoyer}\textsuperscript{*}, {\bf Madian Khabsa}\textsuperscript{*} \\
Meta AI\textsuperscript{*} \qquad Abridge AI\textsuperscript{$\dagger$} \qquad University of California, Los Angeles\textsuperscript{$\mathsection$}
\qquad Reka AI\textsuperscript{$\ddagger$}}

\usepackage{titlesec}
\titlespacing*{\section} {0pt}{1.2ex}{0.4ex}
\titlespacing*{\subsection} {0pt}{1.0ex}{0.4ex}
\begin{document}
\maketitle

\begin{abstract}

We present 
\belebele, a multiple-choice machine reading comprehension (MRC) dataset spanning 122 language variants. Significantly expanding the language coverage of natural language understanding (NLU) benchmarks, this dataset enables the evaluation of text models in high-, medium-, and low-resource languages. Each question is based on a short passage from the \flores-200 dataset and has four multiple-choice answers. The questions were carefully curated to discriminate between models with different levels of general language comprehension. The English dataset on its own proves difficult enough to challenge state-of-the-art language models. Being fully parallel, this dataset enables direct comparison of model performance across all languages. We use this dataset to evaluate the capabilities of multilingual masked language models (MLMs) and large language models (LLMs). We present extensive results and findings, notably that despite significant cross-lingual transfer in English-centric LLMs, much smaller MLMs pretrained on balanced multilingual data still understand far more languages. Overall, \belebele opens up new avenues for evaluating and analyzing the multilingual capabilities of NLP systems.

\end{abstract}

\section{Introduction}

\begin{figure*}
    \includegraphics[width=16cm]{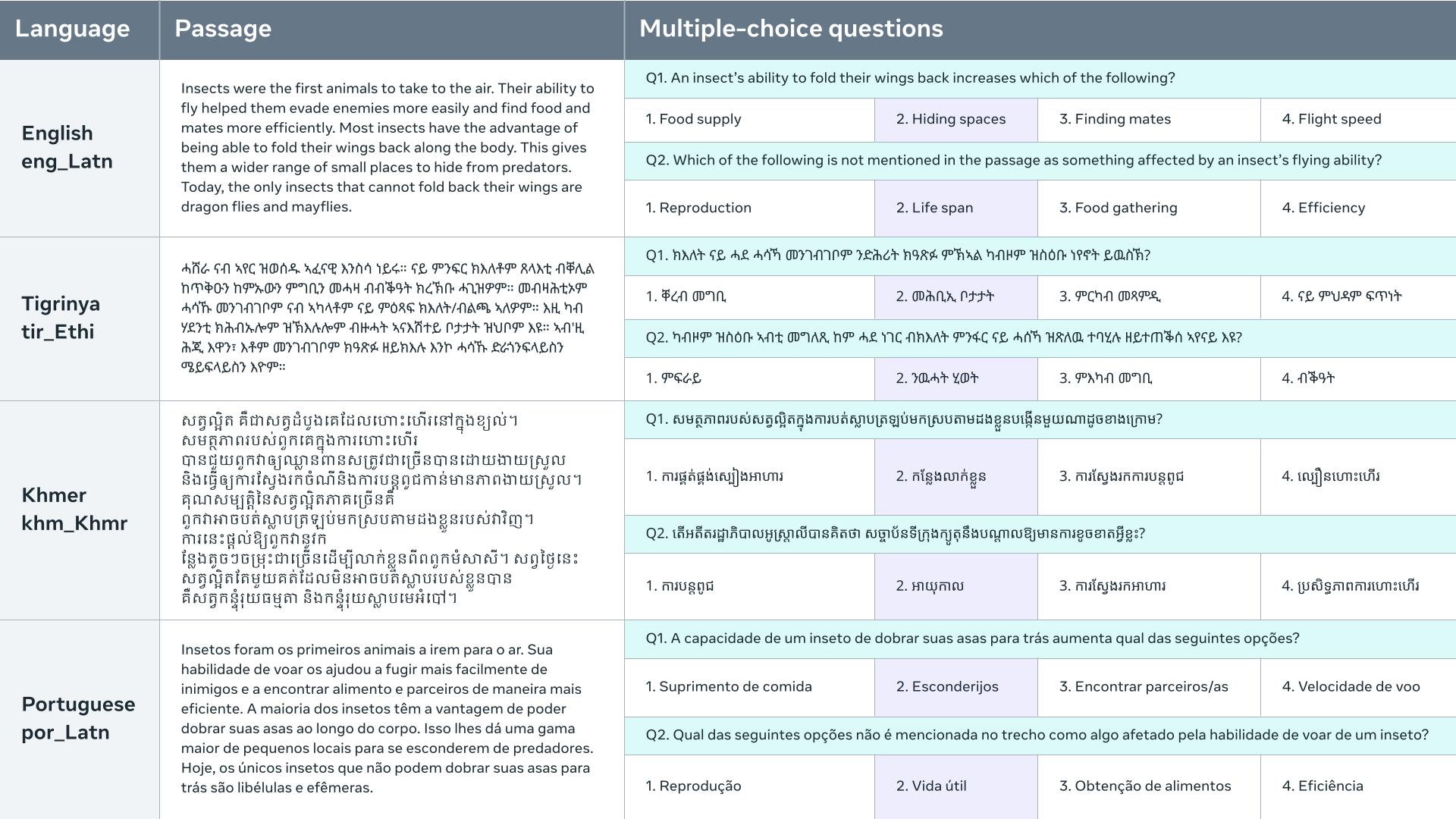}
    \caption{A sample passage from the dataset in 4 different languages, displayed alongside its two questions.}
    \label{fig:samples}
\end{figure*}

The absence of high-quality, parallel evaluation benchmarks is a major obstacle in assessing the text comprehension capabilities of multilingual models. NLP datasets with high language coverage do exist, such as \flores-200 \cite{nllb-2022-no}, but they primarily focus on machine translation. Popular multilingual evaluation benchmarks, such as multilingual question answering \cite{lewis-etal-2020-mlqa,clark-etal-2020-tydi}, natural language inference (NLI) \cite{conneau-etal-2018-xnli}, and summarization \citep{ladhak-etal-2020-wikilingua,hasan-etal-2021-xl}, altogether only cover around 30 languages. 
And while understanding and generative text services are used across the globe in 100+ languages, the lack of labeled data provides a major obstacle to building functional systems in most languages.

Simultaneously, large language models (LLMs) have become increasingly popular. Certain LLMs, like BLOOM \cite{scao2022bloom}, are trained on multilingual data and tout their innate multilingual capabilities. Others like \textsc{GPT-3} \cite{brown2020language} and \llama \cite{touvron2023llama} have demonstrated multilingual competence despite their training data being predominantly in English. Even so, LLMs benefit from pretraining data that is linguistically diverse, intentionally or not, as well as from cross-lingual transfer \cite{zoph-etal-2016-transfer, artetxe-etal-2020-cross, muller-etal-2021-first}. But how multilingual are these models really? Beyond LLMs, significant scientific progress needs to be made before NLP systems can be built effectively and efficiently in low-resource languages. 
Many modeling techniques are being presented as language-\emph{agnostic} but have only truly been evaluated in a small number of languages \cite{Bender2011OnAA}, risking not being applicable to diverse typologically phenomena \cite{bender-2009-linguistically}. We believe that large-scale, parallel, and discriminative datasets are crucial for studying the multilingual capabilities of such models and understanding how the technological disparity between high- and low-resource languages is evolving. 

In this paper, we present a fundamental natural language understanding benchmark to evaluate language models across 122 language variants from around the world\footnote{Download at \href{https://github.com/facebookresearch/belebele}{github.com/facebookresearch/belebele}}, called \belebele\footnote{Bambara word meaning "big, large, fat, great".}. The dataset contains 900 unique multiple-choice reading comprehension passages and questions. The questions have been carefully crafted to discriminate between models with varying competence in language comprehension. While the questions do not necessarily require higher levels of knowledge or reasoning, they favor generalizable NLU models and deliberately punish biased models. The English questions on their own present a significant challenge to numerous models, while humans are capable of answering the questions with near-perfect accuracy.

The first of its scale, \belebele is parallel across all languages, facilitating a direct comparison of model performance across all languages. The dataset covers typologically diverse languages across 29 scripts and \nfams language families. Seven languages are included in two separate scripts, resulting in one of the first NLP benchmarks for the romanized variants of Hindi, Urdu, Bengali, Nepali, and Sinhala. We further detail our data collection process and the resulting corpus in Section \ref{dataset}. 

The dataset enables evaluation of mono- and multi-lingual models, but the parallel nature also enables a number of cross-lingual evaluation settings. We evaluate several masked language models (MLMs) after fine-tuning on an English training set as well as with the assistance of machine translation (Translate-Train-All). For LLMs, we evaluate several models using In-Context Learning and also instruction-tuned models via Zero-Shot. We discuss our results in Section \ref{results}. 

\begin{table*}
    \centering
    \small
    \resizebox{\linewidth}{!}{
        \begin{tabular}{l c l c l c}
            \toprule
            \multicolumn{6}{c}{\textbf{\belebele Statistics}} \\
            \multicolumn{2}{c}{Languages} & \multicolumn{2}{c}{Passage statistics} &
            \multicolumn{2}{c}{Question statistics}  \\
            \midrule
            Total Number & 122	&  Distinct Passages & 488 & Distinct Questions & 900\\
            Distinct Languages (ignoring script) & 115 & Questions per passage & 1-2 & Multiple-choice answers (num correct) per question & 4     (1)\\
            Language Families & \nfams & Avg. words per passage (std) & 79.1 (26.2) & Avg. words per question (std) & 12.9 (4.0) \\
            Scripts & 29 & Avg. sentences per passage (std) & 4.1 (1.4) & Avg. words per answer (std) & 4.2 (2.9) \\
            \bottomrule
        \end{tabular}   
        }
    \caption{Language and Text Information for \belebele. Average length statistics are computed on the English split.}
    \label{stats}
\end{table*}

\section{Background} \label{background}

\subsection{Cross-Lingual Evaluation Benchmarks}
There are several datasets for NLU that are parallel across numerous languages and enable monolingual, multilingual, or cross-lingual evaluation.  These include \textsc{XNLI} \cite{conneau-etal-2018-xnli},  \textsc{XQuAD} \cite{artetxe-etal-2020-cross}, and MLQA \cite{lewis-etal-2020-mlqa}. \textsc{Mintaka} \cite{sen-etal-2022-mintaka} is designed with LLMs in mind, presenting a more difficult QA task in 9 languages. Beyond QA, \textsc{XL-Sum} \cite{hasan-etal-2021-xl} is an analogous dataset in the domain of abstractive summarization. However, all these datasets together cover under 30 languages, most of which are high- or medium- resource. MASSIVE \cite{fitzgerald-etal-2023-massive} is a large NLU dataset covering 51 languages, but in the domain of spoken conversational agents. NER \cite{pan-etal-2017-cross} has extensive language coverage and \textsc{TyDiQA} \cite{clark-etal-2020-tydi} is a popular multilingual benchmark but neither are parallel. 

Our work undertakes the challenge of expanding existing cross-lingual evaluations to 122 languages, many of which currently lack any NLU benchmark at all.

\subsection{Non-English Machine Reading Comprehension}
While the question-answering portion varies, machine reading comprehension (MRC) tasks are defined by the closed-book passage provided to answer each question. Of course, a big majority of MRC datasets are in English, such as \textsc{TriviaQA} \cite{joshi-etal-2017-triviaqa} and the \textsc{bAbI} tasks \cite{DBLP:journals/corr/WestonBCM15}.

However, the need for MRC datasets for other languages has led to a proliferation of monolingual closed-book MRC datasets in recent years \cite{mozannar-etal-2019-neural, hardalov-etal-2019-beyond, dhoffschmidt-etal-2020-fquad, moller-etal-2021-germanquad, anuranjana-hindirc, gupta-etal-2018-mmqa, Croce2018NeuralLF, Efimov_2020, shavrina-etal-2020-russiansuperglue, sun-tibetan}. Most were created using translation and so are parallel with an English dataset, often \textsc{SQuAD} \cite{rajpurkar-etal-2016-squad}. However, \belebele aims to cover these languages and more in one consistent dataset.

\subsection{Multiple Choice QA}
Compared to extractive QA, multiple-choice is a less common form of MRC datasets. Some, like RACE \cite{lai-etal-2017-race}, are made from exam questions for English learners, while others were built specifically for NLU systems, like MCTest \cite{richardson-etal-2013-mctest} and MultiRC \cite{khashabi-etal-2018-looking}. While most are intended to be closed-book, \textsc{SciQ} \cite{welbl-etal-2017-crowdsourcing} and \textsc{OpenBookQA} \cite{mihaylov-etal-2018-suit} require open information retrieval.
Others, like COPA \cite{copa-roemmele}, SWAG \cite{zellers-etal-2018-swag}, and \textsc{ReClor} \cite{Yu2020ReClorAR}, require higher-level commonsense reasoning to answer. 
For multilingual systems, EXAMS \cite{hardalov-etal-2020-exams} is a parallel multiple-choice QA dataset covering 28 languages. However, no passages are provided and answering questions requires cross-lingual knowledge transfer and reasoning.

\subsection{\flores-200}
The passages in the \belebele corpus are directly sourced from the \flores-200 Machine Translation Benchmark \cite{goyal-etal-2022-flores, nllb-2022-no}. The parallel dataset was constructed by sourcing English passages from Wikinews, Wikijunior, and WikiVoyage. The translations were performed by native speakers with high English fluency and translation experience. Translators were instructed to maintain informative and standardized content while handling named entities, abbreviations, idioms, and pronouns appropriately.

\begin{figure*}
    \includegraphics[width=16cm]{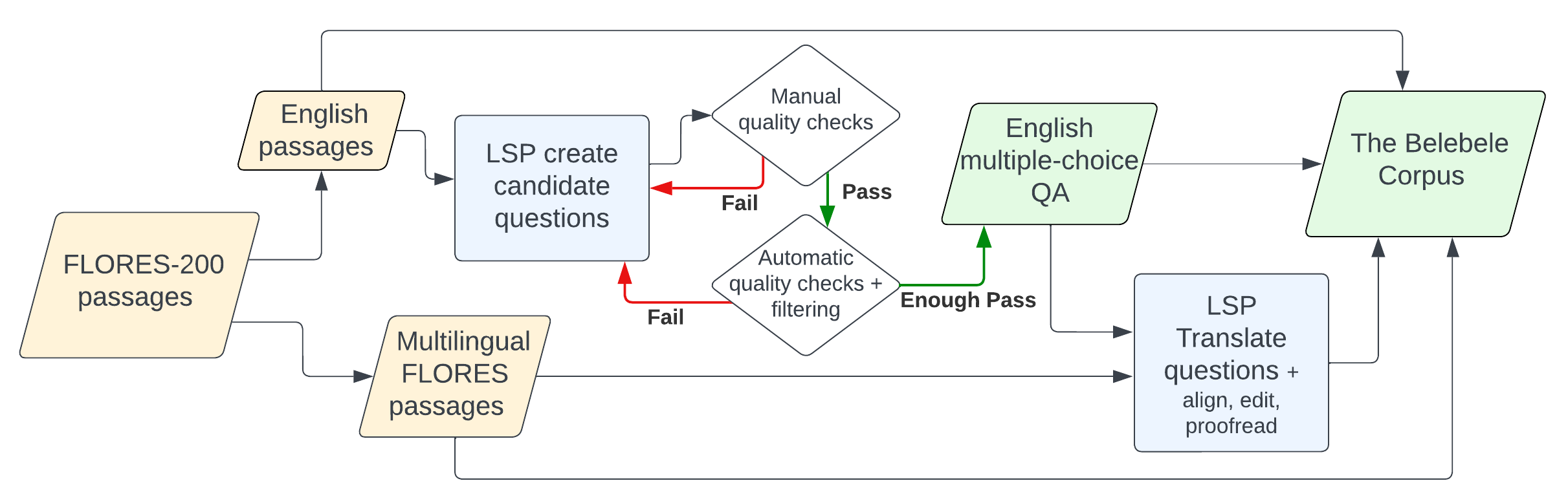}
    \caption{Illustration of the dataset creation process with Language Service Provider (LSP) starting from \flores}
    \label{fig:flowchart}
\end{figure*}

\section{The \belebele Dataset} \label{dataset}

We opted to create multiple-choice questions and answers in English and then translate, as opposed to creating resources natively in each language. Many of the advantages to this approach outlined in \citet{conneau-etal-2018-xnli} remain. Most importantly, this leads to significantly more similar sets of samples across languages, enabling direct score comparison. The process for creating the dataset is summarized in Figure~\ref{fig:flowchart}.

\subsection{Creation of Multiple Choice Questions \& Answers} \label{creation}

To create the \belebele dataset, we first construct a question-answering dataset in English.

Amongst machine reading comprehension tasks, we select multiple-choice questions (MCQs) because it would lead to the fairest evaluation across languages. Related tasks, such as span extraction, are more sensitive to morphological differences, making scaling to many languages difficult \cite{lewis-etal-2020-mlqa}. In addition, MCQs enable us to better center the questions on information explicitly stated in the passage, as yes/no or entailment (NLI) questions can be easier to answer with external knowledge held in pretrained models. In order for the questions to discriminate solely between different levels of language comprehension, we intentionally create questions that do not require higher levels of information processing, such as multi-hop or commonsense reasoning.

Constructing high quality MCQs depends most importantly on creating strong negatives that are neither obviously wrong nor possibly correct \cite{agarwal-mannem-2011-automatic, richardson-etal-2013-mctest}. We do not want the dataset to be easy enough for biased models (e.g. models that use shortcuts or pattern-match) \cite{boyd-graber-borschinger-2020-question}. In setting up this annotation, we consider the protocols proposed in \citet{bowman-etal-2020-new} and the warnings from \citet{malaviya-etal-2022-cascading}. We implement an iterative procedure with the Language Service Provider (LSP) for this involved data collection task, similar to that from \citet{nangia-etal-2021-ingredients}. We engaged in 5 total iterations, providing and receiving feedback in each. Annotators were instructed on the similarities and differences on how ML models approach QA datasets versus humans, which we felt substantially improved the quality of the data.

Our final guidelines include both important points such as having the correct response being unambiguous, as well as particularized rules such as \emph{no double negatives} \cite{mihaylov-etal-2018-suit}. For each rule we provided annotators with a good and bad example to illustrate. An abridged version of our guidelines can be found in the Appendix \ref{guidelines}.

\begin{table*}
    \centering

    \footnotesize
    \begin{tabular}{c c c|c c c c c}
        \toprule
        \textbf{Model} & \textbf{Size/Variant} & \textbf{Vocab size} & \textbf{AVG} & \textbf{\% $\geq$ 50} & \textbf{\% $\geq$ 70} & \textbf{eng\_Latn} & \textbf{non-Eng AVG} \\
        \midrule
        \multicolumn{8}{l}{\emph{5-Shot In-Context Learning (examples in English)}}  \\
        \midrule
        \llama 1 & 7B & 32K & 27.7 & 0.0\% & 0.0\% & 37.3 & 27.6 \\ 
        \llama 1 & 13B & 32K & 30.4 & 0.8\% & 0.0\% & 53.3 & 30.2 \\ 
        \llama 1 & 30B & 32K & 36.2 & 18.0\% & 0.8\% & 73.1 & 35.9 \\ 
        \llama 1 & 70B & 32K & 40.9 & 25.4\% & 12.3\% & 82.5 & 40.5 \\ 
        \llama 2 base & 70B & 32K & \textbf{48.0} & \textbf{38.5\%} & \textbf{26.2\%} & \textbf{90.9} & \textbf{47.7} \\ 
        \textsc{Falcon} & 40B & 65K & 37.3 & 16.4\% & 1.6\% & 77.2 & 36.9 \\
        \midrule
        \multicolumn{8}{l}{\emph{Zero-Shot for Instructed Models (English instructions)}}  \\
        \midrule
        BLOOMZ** & 7.1B & 251K & 43.2 & 28.7\% & 9.0\% & 79.6 & 42.9 \\
        \llamatwochat & 7B & 32K & 34.4 & 4.1\% & 0.0\% & 58.6 & 34.1 \\
        \llamatwochat & 70B & 32K & 41.5 & 27.0\% & 2.5\% & 78.8 & 41.2 \\ 
        \gptturbo & unk & 100K & \textbf{51.1} & \textbf{44.2\%} & \textbf{29.2\%} & \textbf{87.7} & \textbf{50.7} \\ 
        \midrule
        \multicolumn{8}{l}{\emph{Full Finetuning in English}}  \\
        \midrule
        XLM-R & large (550M) & 250K & 54.0 & 64.8\% & 15.6\% & 76.2 & 53.8 \\ 
        XLM-V & large (1.2B) & 902K & 55.6 & \textbf{69.7\%} & 21.2\% & 76.2 & 54.9 \\ 
        \textsc{InfoXLM} & large (550M) & 250K & \textbf{56.2} & 67.2\% & \textbf{28.7\%} & \textbf{79.3} & \textbf{56.0} \\ 
        \midrule
        \multicolumn{8}{l}{\emph{Translate-Train-All}}  \\
        \midrule
        XLM-R & large (550M) & 250K & 58.9 & 69.7\% & 36.1\% & 78.7 & 58.8 \\ 
        XLM-V & large (1.2B) & 902K & \textbf{60.2} & \textbf{76.2\%} & 32.8\% & 77.8 & \textbf{60.1} \\ 
        \textsc{InfoXLM} & large (550M) & 250K & 60.0 & 70.5\% & \textbf{36.9\%} & \textbf{81.2} & 59.8 \\ 
        \bottomrule
    \end{tabular}
    \caption{Summary of results on \belebele across models and evaluation settings. \% $\geq$ 50/70 refers to the proportion of the 122 languages for which a given model performs above 50/70\%. We additionally report \llamatwochat zero-shot results leveraging translation in Table \ref{table:translatetest}. ** In this table, we present results from BLOOMZ despite it being finetuned for translation with \flores data, presenting an unfair advantage for the Zero-Shot setting (See our Ethics Statement).
    }
    \label{table:summary}
    
\end{table*}

\subsection{Quality Assurance}

At each iteration, we evaluate whether or not returned samples satisfy the minimum quality bar through a mix of manual inspection and automatic inspection.
At every step, we manually verified a sample of questions to understand how well the annotators were on the same page with us about the guidelines. While time consuming, manual verification was the most assured way to provide tangible feedback to the annotators, notably on the difficulty of the questions created. As we progressively aligned with annotators, we were required to look over more samples to provide feedback.

To complement the manual inspection of a subset of questions, we use programmatic methods to evaluate all questions from a statistical perspective. Based on the findings in \citet{malaviya-etal-2022-cascading}, we create low-level features to identify overly easy questions or low-effort strategies employed by annotators. For example, we evaluate the lexical overlap between different combinations of the texts associated with a question to evaluate whether the question is answerable by a biased model. This allows us to see if the question can be answered without the passage, without the question, or with only one sentence in the passage. We also identified patterns associated with heuristic solvability, such as the wrong answers less frequently being extracted from the passage. We detail these features in Appendix~\ref{lexical_features}.

These low-level features allow us to (1) determine whether an annotation iteration was up to par, (2) filter out questions that failed these heuristic checks (for the final iteration, about 20\% were filtered out), and (3) compare to other MCQ datasets. We run statistical t-tests to ensure the distribution of these features for correct answers is no different than for wrong answers. In comparison to \textsc{MCTest} which largely fails this t-test (p-value < 0.01), our final collection has p-value 0.81. We also train a logistic regression model to answer using only bag-of-word representations and find that the best the naïve model could achieve was an accuracy of 0.28 on our 900 questions. This is just better than random (0.25) and much lower than what was achieved on \textsc{MCTest}, 0.44.

\subsection{Translating the Corpus} \label{translating}
\belebele was created end-to-end without the use of machine translation technology, relying solely on experts fluent in English and the target language.

For all languages included in the corpus, the context passages were taken directly from the \flores-200 dataset, with the exception of Hindi, Bengali, Urdu, Nepali, and Sinhala in the Latin script. While the romanized variant of these 5 Indo-Aryan languages is very prevalent on the modern Internet, their romanization is not included in \flores-200. We thus had annotators transliterate from the native to Latin script with the support of IndicXlit \cite{aksharantar}. As a result, much like Modern Standard Arabic, these languages are present in two forms in the corpus.

In order for the questions and answers to properly pair the translated \flores passages, the latter was provided for the annotators. We specifically instructed annotators to align potentially ambiguous translations with the original passages. While \citet{clark-etal-2020-tydi} warns that this forced alignment could increase `translationese', it is necessary to ensure equivalent question difficulty across languages. The modifications to the translation guidelines can be found in Appendix \ref{specifications}. All translations were proofread and edited by an additional annotator.

\subsection{English Training Data} 
\belebele is intended to be used as a test set, and not for training. Therefore, for models that require additional task finetuning, we instead assemble a training set consisting of samples from English multiple-choice QA datasets (See Appendix \ref{trainset}).

\subsection{The \belebele Dataset in Summary}

\belebele contains 900 questions, each with 4 multiple-choice answers and one correct answer. Most passages have two associated questions, but some have only one. In total, there are 488 distinct passages, none belonging to the hidden \flores test set. Parallel across 122 languages, the corpus contains a total of 109,800 rows. Amongst the language varieties, there are 29 unique scripts and 27 language families represented (see Figure~\ref{table:languages}). Some text and language statistics are displayed in Table~\ref{stats} and we display a sample passage in four languages in Figure~\ref{fig:samples}.

Because of the careful annotation procedure and quality checks, the MCQs discriminate text comprehension competence. It often includes paraphrasing and strong negatives in order to elude simple pattern-matching models. Questions often additionally require understanding multiple sentences. However, answering does not require presumptions or external knowledge as is required in more difficult reasoning datasets. For example, Q1 in Figure~\ref{fig:samples} is unambiguous. \emph{Food}, \emph{mates}, and \emph{flying} are all mentioned in the passage, but a careful read reveals the wings folding back is only associated with \emph{hiding spaces}. To confidently rule out the other candidate answers, it is required to understand three sentences. In general, we find all questions to be answerable by humans fluent in the target language, but not without focused reading (see Section \ref{difficult}).

As can be seen in Figure~\ref{fig:samples}, the passages, questions, and answers are aligned in semantic meaning and formality. \belebele therefore poses an equivalent challenge in all languages. It also enables models with cross-lingual alignment in the semantic representation space to answer questions when passage, question, and answers are swapped to different languages. Since \flores includes passages in 83 additional languages, we can even evaluate reading comprehension in these languages by asking the questions in English.

\section{Experiments} \label{experiments}

Thanks to \textsc{Belebele}, we are able to evaluate numerous models and establish baseline performances across 122 language variants. We compare performance between popular multilingual MLMs and LLMs in several settings. For all, accuracy is the central metric. With 4 candidate answers for each question, the expected accuracy for sequence classification models that guess randomly is 0.25.\footnote{Note: For sequence-to-sequence models (e.g. instructed models) that are evaluated in exact-match scenarios, this lower-bound does not hold.}

\begin{figure*}
    \noindent
    \includegraphics[width=16cm]{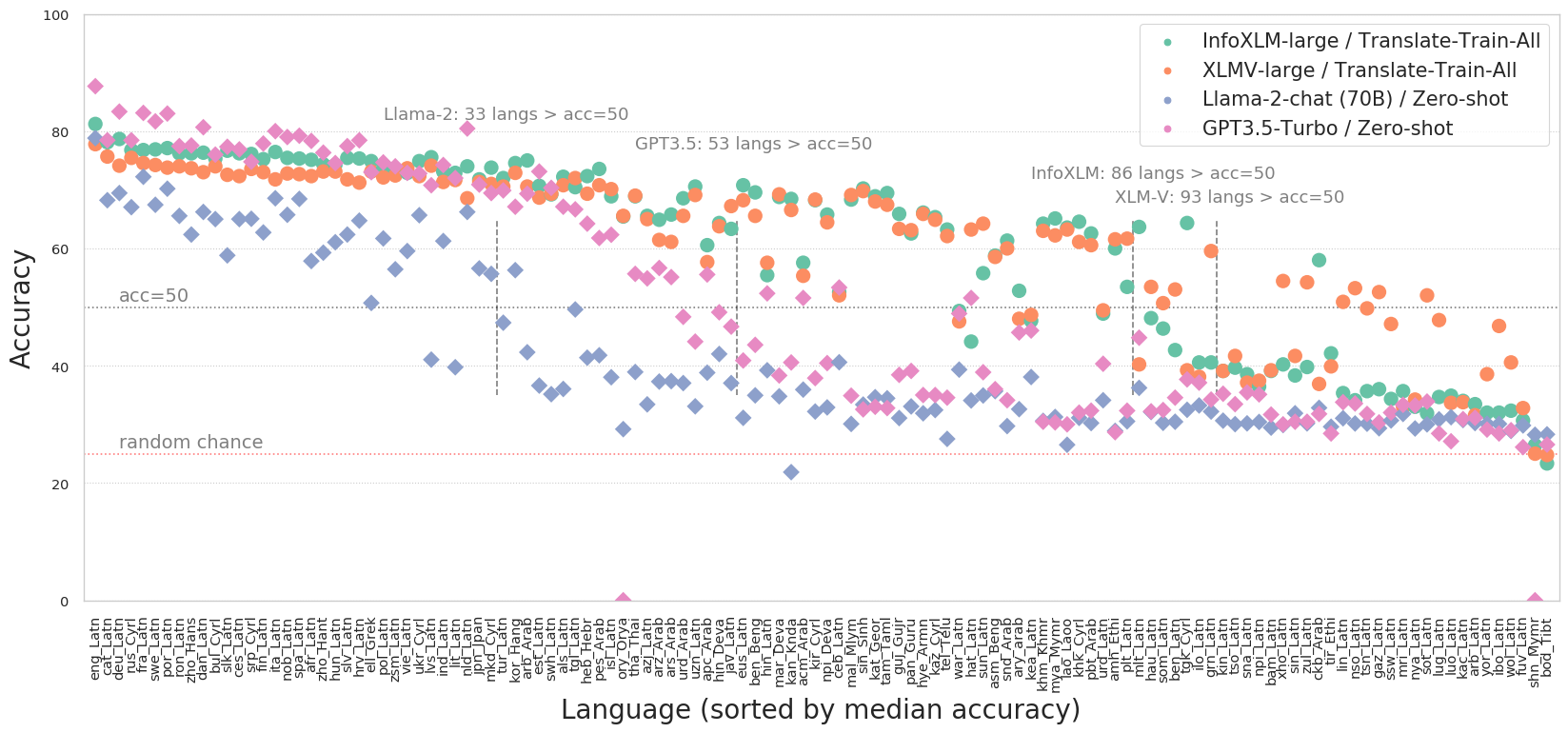}
    \caption{\belebele Results in 122 languages.  We compare four models in two settings and see the difference between intentionally multilingual models and models with English-centric data.  \gptturbo performs the best on the top 20 languages, but after 40-50, its performance falls far behind \textsc{InfoXLM} and XLM-V. Similarly, \textsc{InfoXLM} outperforms XLM-V in the first 40 languages, but XLM-V proves more capable on the long tail of languages. Note that the language order can change the plot considerably, here we choose median accuracy. }
    \label{results122}
\end{figure*}

\subsection{Evaluated Models}
\label{sec:evaluated_models}

\paragraph{Masked Language Models (MLMs)}
We evaluate three different models, \textsc{XLM-V} \cite{liang2023xlm}, \textsc{InfoXLM} \cite{chi-etal-2021-infoxlm}, %
and \textsc{XLM-R} \cite{conneau-etal-2020-unsupervised}. All the evaluated MLMs have been pretrained on intentionally multilingual corpi inclusive of about 100 languages. The pretraining data in high-resource languages is typically down-sampled while low-resource languages are up-sampled in order to favor multilingual performance \citep{conneau-etal-2020-unsupervised}. In addition, all their subword tokenizers \citep{kudo-richardson-2018-sentencepiece} are trained on multilingual corpora, making them better suited for multilingual text.

\paragraph{Large Language Models}
We evaluate \gptturbo, \falcon, and \llama (1 and 2). \gptturbo is a model optimized for chat based on GPT-3 \citep{brown2020language} available through OpenAI APIs\footnote{\url{https://platform.openai.com/docs/models}}. Limited details have been disclosed about the pretraining and fine-tuning data.\footnote{Our analyses rely on the unverified assumption that \gptturbo was not trained on \flores. See Limitations} \llama~1 \citep{touvron2023llama} is a collection of decoder-only transformers models trained on 1T (for 7B, 13B) or 1.4T (for 30B, 65B) tokens of publicly available online data, while  \llama~2 \citep{touvron2023llama2} is pretrained on about 2T. We evaluate all four pretrained checkpoints for \llama~1. We evaluate both \llama~2 70B's pretrained version and its chat version instruction-fine-tuned for safe dialog purposes (a.k.a. \llamatwochat). We also evaluate \falcon 40B, which was pretrained on 1T extensively filtered web-crawled samples \cite{penedo2023refinedweb}.

While \llama 1 was reportedly trained in 20 languages with Latin and Cyrillic scripts, non-English text accounts for less than 4.5\% of the pretraining corpus \citep{touvron2023llama}. 
\llama 2 pretraining data is made of 89.7\% of English data, 8.4\% unidentified, and a tiny 1.9\% belonging to 26 other languages\footnote{See Table 10 in \citet{touvron2023llama2} for a full list of the identified languages} \citep{touvron2023llama2}. Both series use the same BPE-based tokenizers \citep{kudo-richardson-2018-sentencepiece}. Splitting unicode characters into bytes also helps \llama avoid out-of-vocabulary errors.

In Table~\ref{table:summary}, we present results for BLOOMZ-7B~\cite{muennighoff-etal-2023-crosslingual}, which was both pretrained and instruction finetuned on heavily multilingual data and has significant vocabulary capacity. However, we do not discuss it in our analyses, as it was fine-tuned for translation on \flores-200, undermining fair assessment.

\subsection{Evaluation Settings} \label{evalsettings}
More specifics are provided in Appendix \ref{evaldetails}.
\paragraph{Full Model Fine-tuning}

For evaluating MLMs, we add a multiple-choice classification head and fine-tune the entire model. We finetune in two settings, (1) in English and evaluate zero-shot cross-lingual transfer and (2) on machine-translated samples of the training set to all the target languages and evaluate each language (\emph{Translate-Train-All}).

\paragraph{Five-Shot In-Context Learning}

We evaluate the pretrained \llama 1 and 2 as well as \falcon~40B in the five-shots setting. Examples are sampled from the English training set and prompted to the model. For prediction, we pick the answer with highest probability and report the average cumulative score over 3 runs.

\paragraph{Zero-Shot Evaluation}

We evaluate GPT3.5 and \llamatwochat (70B) in Zero-Shot by describing the task with natural language instructions (in English). We present the passage, question, and four possible answers, and instruct the model to provide the letter of the answer. We post-process answers and accept multiple formats.\footnote{Code provided at \href{https://github.com/facebookresearch/belebele}{
github.com/facebookresearch/belebele}}

In addition, we prompt \llamatwochat with instructions that are machine translated to the target language from English. Conversely, we evaluate machine-translating the passages, questions, and answers back to English and prompting them to the model (\emph{Translate-Test}). This setting allows us to compare in-language comprehension to the popular approach of cascading with machine translation.

\section{Results} \label{results}

We provide summarized results in Table~\ref{table:summary} and detailed results in Appendix \ref{details}.

\subsection{How difficult is \belebele?} \label{difficult}

As discussed in Section \ref{dataset}, the questions in \belebele are intentionally difficult. While the primary challenge of this dataset is its multilinguality, we see that empirically, the English questions are able to shed light on the varying NLU capabilites of models. With full finetuning, we achieved a maximum accuracy of $71.7$ in English with \textsc{RoBERTa}-base model, significantly less than the $90.9$ achieved by \llama~2 70B in five-shot. Between \llama~1 variants, we see a wide range of results, with the 7B model only achieving $37.3$. So while the primary difficulty of \belebele is its multilinguality, we see a wide range of performance between different model types and sizes.

In addition, all models evaluated comfortably underperform humans. To establish human performance, 4 authors each randomly sampled around 30 English MCQs and answered in a blind test, achieving mean $97.6$ accuracy\footnote{95\% CI for all 900 questions = $[93.1, 99.5]$}. This is much higher than any of the models evaluated, implying the task presents a particular challenge for models and that there is room to improve. For comparison, \citet{nangia-bowman-2019-human} conservatively estimate human performance to be 92.8 on the English split of XNLI (i.e. MNLI \citep{williams-etal-2018-broad}).

When comparing model performance with XNLI, we find very high correlation. In the Translate-Train-All setting, XLM-V, \textsc{InfoXLM}, and XLM-R all perform about 10 accuracy points lower on \belebele than on XNLI Translate-Train\footnote{In traditional Translate-Train, the model is finetuned on translated training inputs for each language \emph{individually}.} reported in \citet{liang2023xlm} and \citet{chi-etal-2021-infoxlm}. Still, across all 15 languages and three models, we find a score correlation of $r=0.85$. 

\begin{table*}
    \centering

    \footnotesize
    \begin{tabular}{c c c|c c c c}
        \toprule
        \textbf{Model} & \textbf{Variant} & \textbf{Eval Setting} & \textbf{AVG} & \textbf{\% $\geq$ 50} & \textbf{\% $\geq$ 70} &
        \textbf{eng\_Latn} \\
        \midrule
        \multicolumn{7}{l}{\emph{Translate-Test (English) on 91 non-English languages in Zero-Shot}}  \\
        \midrule
        \llamatwochat & 70B & Translate-Test & \textbf{57.1} & \textbf{78.0\%} & 2.2\% & 78.8  \\ 
        \llamatwochat & 70B & In-Language & 44.1 & 35.2\% & 2.2\% & 78.8  \\
        \midrule
        \multicolumn{7}{l}{\emph{Translated Instructions in 89 non-English languages Zero-Shot}}  \\
        \midrule
        \llamatwochat & 70B & In-Language Translated Instructions & 38.7 & 36.0\% &  \textbf{7.9\%} & 78.8  \\ 
        \llamatwochat & 70B & English Instructions & \textbf{44.9} &  \textbf{37.1\%} & 3.4\% & 78.8  \\
        \bottomrule
    \end{tabular}
    \caption{Results of \llamatwochat in zero-shot in two machine translation-based evaluation settings; Translate-Test (passages, questions, answers translated back to English) and evaluations with the English instructions translated to the target language. The traditional setting on the same languages is provided for comparison. Note that the summary statistics differ from Table~\ref{table:summary} because this is for a subset of all languages on which we performed these translation-based evaluations.
    }
    \label{table:translatetest}
    
\end{table*}

\subsection{Multilingual Generalization of MLMs and LLMs on \belebele}

Schematically, the performance of a language model in a given language is related to two key factors. 
(i)~First, the amount of pretraining data in the target language. As predicted by the scaling laws \citep{kaplan2020scaling}, performance in a language increases monotonically with the amount of pretraining tokens. 
(ii)~Second, the cross-lingual transfer happening between languages in the pretraining data and the target language at inference time \citep{conneau-etal-2020-unsupervised,conneau-etal-2020-emerging}. This transfer is impacted by a combination of typological, script, and lexical similarities between the pretraining languages and the target language \citep{muller-etal-2021-unseen,pmlr-v203-muller23a}. These two factors are hard to disentangle due to the scale (up to $\sim$1T tokens) and the potential language leaks of large-scale pretraining corpora \cite{kreutzer-etal-2022-quality}. 
Thanks to \belebele's quality and scale, we provide detailed evidence of both impacting the multilingual generalization of the models.

\paragraph{Impact of Pretraining Language Distribution}

One of the key differences between the MLMs and LLMs evaluated is their pretraining data distribution and parameter size, explaining the large performance differences between them. For instance, \llama~2 largely outperforms XLM-R on high-resource languages, but only achieves accuracy 50 on about half the amount of languages as XLM-R (See Table \ref{table:summary}). This difference between the MLMs and LLMs evaluated is illustrated in Figure~\ref{results122}. However, despite this gap, all LLMs evaluated perform surprisingly well on a large number of languages. For instance, \llamatwochat is above 35 accuracy (i.e. 10 above random) for 59 languages. This shows that English-centric LLMs are a promising starting point to build multilingual models.

\paragraph{Machine Translation for Zero-Shot}

Our Translate-Test evaluations show that using machine translation into English strongly outperforms \llamatwochat (70B) performance in the original target language.  Across 91 evaluated languages, only 2 are non-trivially better in-language (German and Italian), compared to 68 (none high-resource) for which translating to English is better, none of which are considered high-resource. Compared to \llamatwochat having zero-shot accuracy above 50\% for 33 languages, it has 71 in
Translate-Test (see Appendix \ref{tt_appendix}).

\setlength{\belowcaptionskip}{-8pt}
\captionsetup[figure]{skip=-10pt}
\begin{figure*}
    \noindent
    \includegraphics[width=16cm]{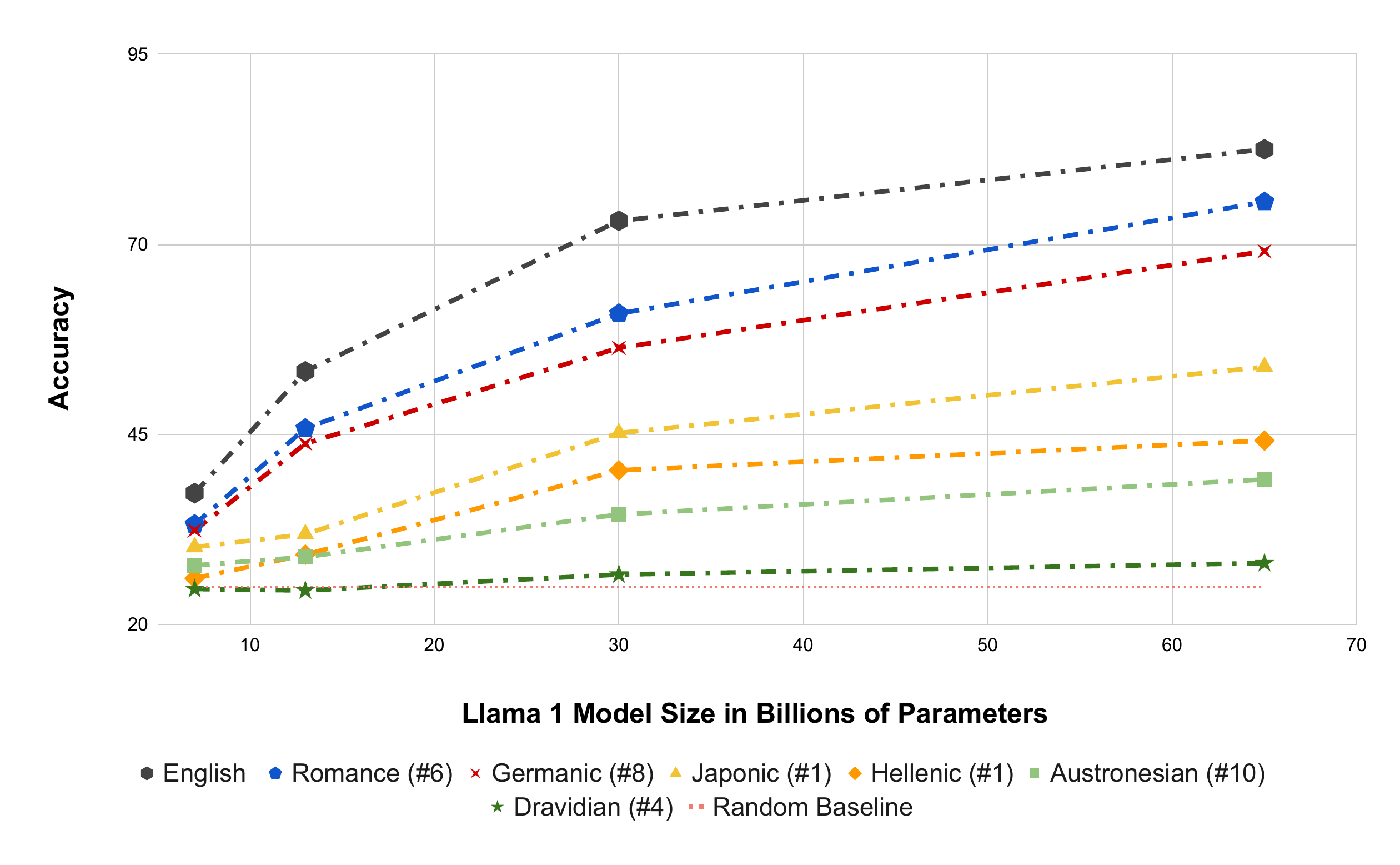}
    \caption{Impact of Models' scale (from 7B to 65B parameters  of \llama 1) on the performance on \belebele for 6 language families and English. The number of languages in a given family is indicated as (\#N). \llama 1 is evaluated in the 5-shot settings with examples from the training data in English. Scores are averaged over 3 runs.}
    \label{llama_scaling}
\end{figure*}

In addition, we evaluate machine-translating the task instructions to the target language. For around 25 languages, the translated instructions were not well understood (i.e. accuracy less than random), correlating strongly with already low-score languages. For the rest, the performance relative to using English instructions is mixed, though languages that already scored highly had the largest accuracy boost from in-language instructions. While machine-translating instructions are less effective than quality in-language instructions, these results do not suggest that the use of English instructions is the primary reason for why performance on non-English languages lags English so significantly.

\paragraph{Impact of Sub-Word Tokenization}
We reaffirm a correlation between increasing vocabulary size and performance on lower resource languages \cite{liang2023xlm}. XLM-V has a massive 900k-token vocabulary that allocates capacity for each individual language and de-emphasizes token sharing between languages. XLM-V outperforms XLM-R and \textsc{InfoXLM} (250k vocabulary size) on low-resource languages even though they all have the same architecture and are trained on the same dataset (CC-100). \gptturbo (100k vocabulary size),\footnote{According to https://github.com/openai/tiktoken.} \falcon (65k vocabulary size), and \llama 2 (32k vocabulary size) all fall off abruptly for medium- and low- resource languages. Larger vocabulary size may explain why \falcon 40B performs equivalent to \llama~1 30B despite having been pretrained on fewer non-English tokens.

\paragraph{Scaling effect on Multilingual Generalization} \label{scaling}

We report in Figure~\ref{llama_scaling} the impact of model sizes on performance on the \belebele benchmark across six language families and English. We find that scale is critical for \llama 1 to perform reading comprehension as the 7B checkpoint performs slightly above chance in English even. As the parameter size increases, performances across the board increases significantly. 
Only the 30B and 65B checkpoints are able to perform non-trivially in language families not reported to be in the pretraining corpus (Japanese and Greek). However, unlike other lan- guage families such as Romance and Germanic, the performance becomes non-trivial only with the 30B and 65B checkpoints. Results like this suggest that generalizing to distant languages after English-centered pretraining requires more parameters.

\paragraph{Impact of Script}
Comparing the Romanized versions with the original scripts for Hindi, Urdu, Bengali, Sinhala, Nepali, and Modern Standard Arabic, we find that all models except \falcon perform stronger in the native script  than in the Latin script (see Appendix~\ref{scripts_appendix}). However, the native scripts are allegedly not present in the pretraining data for \llama 2 and \falcon. For the Indo-Aryan languages, we hypothesized cross-lingual transfer would be higher in the Latin variant since the tokenization will be more suitable and there is opportunity for shared subwords (anchor points) \cite{conneau-etal-2020-emerging, muller2020multilingual, pfeiffer-etal-2021-unks, muller-etal-2021-unseen}. However, this only seems to be the case for \falcon. The results generally suggest the models were pretrained on significant samples in the native script (perhaps due to code-switching or poor language identification).   %

\section{Conclusion} \label{conclusion}

A fundamental limitation to conducting sound evaluations of the capabilities of language models in low-, or even moderate-, resource languages is the availability of annotated benchmarks. This paper presents a massive dataset, \belebele, consisting of passages and multiple-choice questions evaluating reading comprehension in 122 languages. This benchmark enables critical evaluation of reading comprehension capabilities of LLMs in English and top languages. In addition, the dataset is the first of its kind in many medium- and low-resource languages, enabling unprecedented insight into the multilingual capabilities of language models. We present results from a number of popular MLMs and LLMs in different evaluation settings and find that while large vocabulary size and balanced pretraining data correlates with highest model performance on medium- and low-resource languages, even English-centric LLMs can go a long way and generalize to over 30 languages. 

For future work, we hope the many evaluations and experiments now possible will allow for deeper dives into current language models.
\belebele can also complement investigations into specific model capabilities (e.g. reasoning), leading to a broader understanding of the relationship between such abilities and multilinguality.
As a result, we believe \belebele will soon unveil further insights that contribute to the development of NLP systems beyond high-resource languages.

\section*{Limitations}

\paragraph{Pretraining Documentation} Our model analyses are limited by inconsistent documentation on the composition of pretraining corpora used. We present results from BLOOMZ \cite{muennighoff-etal-2023-crosslingual} only in very limited fashion because its documentation showed it was fine-tuned for translation on \flores-200. However, as alluded to in Section \ref{sec:evaluated_models}, we present \gptturbo results on \belebele even though we cannot verify its pretraining or finetuning data. Because of this, comparing results on \gptturbo to other models may be unfair given the lack of transparency on training data. To enable further understanding of the interaction of multilingual text during training, we point to two critical research directions to enable progress in the field. First, (i)~better language identification systems: popular language identification models are trained on a restricted number of languages and domains and only work at the sentence level \citep{bojanowski-etal-2017-enriching}, limiting their abilities to track languages in code-switched data and embedded text.  Second, (ii)~we encourage LLM developers to improve reporting on pretraining language distribution. This is necessary for the research community to understand the cross-lingual transfer capabilities of LLMs and to improve NLP system design for low-resource languages.

\paragraph{Errors in \flores} As briefly mentioned in Section \ref{translating}, annotators discovered a few quality issues with the \flores translations (i.e. the original annotations done prior to this work). Some of them are likely due to style/dialect differences between annotators, but many seem to not be. It's rare enough, thanks to the extensive quality-assurance loops implemented by the NLLB team and the LSP. However, over the scale of 122 languages a fair number of issues have arisen, especially in lower-resource languages. Since updating the base \flores dataset is not in scope for this project, we deliberated on each with the LSP to maximize both appropriateness and cross-language consistency of the question/answers translations. 

\paragraph{Translationese} Even with our extensive quality assurance, we warn that "translationese" may change the nature of the task task across languages. In numerous cases, the \emph{perfect} translation does not exist. This may cause accuracy on non-English languages to not be directly comparable to on English itself.

\section*{Ethics Statement} \label{ethics}
\paragraph{Open-Source} Our decision to open-source \belebele may compromise future benchmarking as the samples may get collected into large pretraining corpora, undermining fair comparison. This is especially the case for zero- or few-shot evaluation. Nonetheless, we consciously determine the value of open-sourcing the full dataset to far outweigh these considerations.

\paragraph{English-centrism} \belebele was designed to measure the reading comprehension abilities of NLP systems across 122 languages. We specifically align as much as possible with translation choices made in the creation of \flores. Therefore, by-design the samples collected do not capture language- and culture-specific phenomena such as formality \citep{ersoy-etal-2023-languages}, values \citep{kovavc2023large}, and aboutness \citep{hershcovich-etal-2022-challenges}. While conscious of this Western-centrism, \belebele was designed to prioritize comparability across languages. Following \belebele, building NLP systems inclusive of all cultures and languages will require the release of benchmarks that capture these phenomena.

\bibliography{anthology,custom}
\bibliographystyle{acl_natbib}

\titlespacing*{\section} {0pt}{4ex}{0.4ex}
\titlespacing*{\subsection} {0pt}{6ex}{0.4ex}
\captionsetup[table]{skip=12pt}
\captionsetup[figure]{skip=8pt}
\setlength{\belowcaptionskip}{5pt}
\onecolumn
\appendix
\section{Appendix}

\subsection{Languages and Variants}

\begin{small}
\begin{longtable}{llll}
    \toprule
        \textbf{Language Code} & \textbf{Name in English} & \textbf{Script} & \textbf{Family} \\ \midrule
        acm\_Arab & Mesopotamian Arabic & Arab & Afro-Asiatic \\ %
        afr\_Latn & Afrikaans & Latn & Germanic \\ %
        als\_Latn & Tosk Albanian & Latn & Paleo-Balkanic \\ %
        amh\_Ethi & Amharic & Ethi & Afro-Asiatic \\ %
        apc\_Arab & North Levantine Arabic & Arab & Afro-Asiatic \\ %
        arb\_Arab & Modern Standard Arabic & Arab & Afro-Asiatic \\ %
        arb\_Latn & Modern Standard Arabic (Romanized) & Latn & Afro-Asiatic \\ %
        ars\_Arab & Najdi Arabic & Arab & Afro-Asiatic \\ %
        ary\_arab & Moroccan Arabic & Arab & Afro-Asiatic \\ %
        arz\_Arab & Egyptian Arabic & Arab & Afro-Asiatic \\ %
        asm\_Beng & Assamese & Beng & Indo-Aryan \\ %
        azj\_Latn & North Azerbaijani & Latn & Turkic \\ %
        bam\_Latn & Bambara & Latn & Mande \\ %
        ben\_Beng & Bengali & Beng & Indo-Aryan \\ %
        ben\_Latn & Bengali (Romanized) & Latn & Indo-Aryan \\ %
        bod\_Tibt & Standard Tibetan & Tibt & Sino-Tibetan \\ %
        bul\_Cyrl & Bulgarian & Cyrl & Balto-Slavic \\ %
        cat\_Latn & Catalan & Latn & Romance \\ %
        ceb\_Latn & Cebuano & Latn & Austronesian \\ %
        ces\_Latn & Czech & Latn & Balto-Slavic \\ %
        ckb\_Arab & Central Kurdish & Arab & Iranian \\ %
        dan\_Latn & Danish & Latn & Germanic \\ %
        deu\_Latn & German & Latn & Germanic \\ %
        ell\_Grek & Greek & Grek & Hellenic \\ %
        eng\_Latn & English & Latn & Germanic \\ %
        est\_Latn & Estonian & Latn & Uralic \\ %
        eus\_Latn & Basque & Latn & Basque \\ %
        fin\_Latn & Finnish & Latn & Uralic \\ %
        fra\_Latn & French & Latn & Romance \\ %
        fuv\_Latn & Nigerian Fulfulde & Latn & Atlantic-Congo \\ %
        gaz\_Latn & West Central Oromo & Latn & Afro-Asiatic \\ %
        grn\_Latn & Guarani & Latn & Tupian \\ %
        guj\_Gujr & Gujarati & Gujr & Indo-Aryan \\ %
        hat\_Latn & Haitian Creole & Latn & Atlantic-Congo \\ %
        hau\_Latn & Hausa & Latn & Afro-Asiatic \\ %
        heb\_Hebr & Hebrew & Hebr & Afro-Asiatic \\ %
        hin\_Deva & Hindi & Deva & Indo-Aryan \\ %
        hin\_Latn & Hindi (Romanized) & Latn & Indo-Aryan \\ %
        hrv\_Latn & Croatian & Latn & Balto-Slavic \\ %
        hun\_Latn & Hungarian & Latn & Uralic \\ %
        hye\_Armn & Armenian & Armn & Armenian \\ %
        ibo\_Latn & Igbo & Latn & Atlantic-Congo \\ %
        ilo\_Latn & Ilocano & Latn & Austronesian \\ %
        ind\_Latn & Indonesian & Latn & Austronesian \\ %
        isl\_Latn & Icelandic & Latn & Germanic \\ %
        ita\_Latn & Italian & Latn & Romance \\ %
        jav\_Latn & Javanese & Latn & Austronesian \\ %
        jpn\_Jpan & Japanese & Jpan & Japonic \\ %
        kac\_Latn & Jingpho & Latn & Sino-Tibetan \\ %
        kan\_Knda & Kannada & Knda & Dravidian \\ %
        kat\_Geor & Georgian & Geor & kartvelian \\ %
        kaz\_Cyrl & Kazakh & Cyrl & Turkic \\ %
        kea\_Latn & Kabuverdianu & Latn & Portuguese Creole \\ %
        khk\_Cyrl & Halh Mongolian & Cyrl & Mongolic \\ %
        khm\_Khmr & Khmer & Khmr & Austroasiatic \\ %
        kin\_Latn & Kinyarwanda & Latn & Atlantic-Congo \\ %
        kir\_Cyrl & Kyrgyz & Cyrl & Turkic \\ %
        kor\_Hang & Korean & Hang & Koreanic \\ %
        lao\_Laoo & Lao & Laoo & Kra-Dai \\ %
        lin\_Latn & Lingala & Latn & Atlantic-Congo \\ %
        lit\_Latn & Lithuanian & Latn & Balto-Slavic \\ %
        lug\_Latn & Ganda & Latn & Atlantic-Congo \\ %
        luo\_Latn & Luo & Latn & Nilo-Saharan \\ %
        lvs\_Latn & Standard Latvian & Latn & Balto-Slavic \\ %
        mal\_Mlym & Malayalam & Mlym & Dravidian \\ %
        mar\_Deva & Marathi & Deva & Indo-Aryan \\ %
        mkd\_Cyrl & Macedonian & Cyrl & Balto-Slavic \\ %
        mlt\_Latn & Maltese & Latn & Afro-Asiatic \\ %
        mri\_Latn & Maori & Latn & Austronesian \\ %
        mya\_Mymr & Burmese & Mymr & Sino-Tibetan \\ %
        nld\_Latn & Dutch & Latn & Germanic \\ %
        nob\_Latn & Norwegian Bokmål & Latn & Germanic \\ %
        npi\_Deva & Nepali & Deva & Indo-Aryan \\ %
        npi\_Latn & Nepali (Romanized) & Latn & Indo-Aryan \\ %
        nso\_Latn & Northern Sotho & Latn & Atlantic-Congo \\ %
        nya\_Latn & Nyanja & Latn & Afro-Asiatic \\ %
        ory\_Orya & Odia & Orya & Indo-Aryan \\ %
        pan\_Guru & Eastern Panjabi & Guru & Indo-Aryan \\ %
        pbt\_Arab & Southern Pashto & Arab & Indo-Aryan \\ %
        pes\_Arab & Western Persian & Arab & Iranian \\ %
        plt\_Latn & Plateau Malagasy & Latn & Austronesian \\ %
        pol\_Latn & Polish & Latn & Balto-Slavic \\ %
        por\_Latn & Portuguese & Latn & Romance \\ %
        ron\_Latn & Romanian & Latn & Romance \\ %
        rus\_Cyrl & Russian & Cyrl & Balto-Slavic \\ %
        shn\_Mymr & Shan & Mymr & Kra-Dai \\ %
        sin\_Latn & Sinhala (Romanized) & Latn & Indo-Aryan \\ %
        sin\_Sinh & Sinhala & Sinh & Indo-Aryan \\ %
        slk\_Latn & Slovak & Latn & Balto-Slavic \\ %
        slv\_Latn & Slovenian & Latn & Balto-Slavic \\ %
        sna\_Latn & Shona & Latn & Atlantic-Congo \\ %
        snd\_Arab & Sindhi & Arab & Indo-Aryan \\ %
        som\_Latn & Somali & Latn & Afro-Asiatic \\ %
        sot\_Latn & Southern Sotho & Latn & Atlantic-Congo \\ %
        spa\_Latn & Spanish & Latn & Romance \\ %
        srp\_Cyrl & Serbian & Cyrl & Balto-Slavic \\ %
        ssw\_Latn & Swati & Latn & Atlantic-Congo \\ %
        sun\_Latn & Sundanese & Latn & Austronesian \\ %
        swe\_Latn & Swedish & Latn & Germanic \\ %
        swh\_Latn & Swahili & Latn & Atlantic-Congo \\ %
        tam\_Taml & Tamil & Taml & Dravidian \\ %
        tel\_Telu & Telugu & Telu & Dravidian \\ %
        tgk\_Cyrl & Tajik & Cyrl & Iranian \\ %
        tgl\_Latn & Tagalog & Latn & Austronesian \\ %
        tha\_Thai & Thai & Thai & Kra-Dai \\ %
        tir\_Ethi & Tigrinya & Ethi & Afro-Asiatic \\ %
        tsn\_Latn & Tswana & Latn & Atlantic-Congo \\ %
        tso\_Latn & Tsonga & Latn & Afro-Asiatic \\ %
        tur\_Latn & Turkish & Latn & Turkic \\ %
        ukr\_Cyrl & Ukrainian & Cyrl & Balto-Slavic \\ %
        urd\_Arab & Urdu & Arab & Indo-Aryan \\ %
        urd\_Latn & Urdu (Romanized) & Latn & Indo-Aryan \\ %
        uzn\_Latn & Northern Uzbek & Latn & Turkic \\ %
        vie\_Latn & Vietnamese & Latn & Austroasiatic \\ %
        war\_Latn & Waray & Latn & Austronesian \\ %
        wol\_Latn & Wolof & Latn & Atlantic-Congo \\ %
        xho\_Latn & Xhosa & Latn & Atlantic-Congo \\ %
        yor\_Latn & Yoruba & Latn & Atlantic-Congo \\ %
        zho\_Hans & Chinese (Simplified) & Hans & Sino-Tibetan \\ %
        zho\_Hant & Chinese (Traditional) & Hant & Sino-Tibetan \\ %
        zsm\_Latn & Standard Malay & Latn & Austronesian \\ %
        zul\_Latn & Zulu & Latn & Atlantic-Congo \\ \bottomrule
        \caption{The 122 Languages \& Scripts in \belebele.}
        \label{table:languages}
    \end{longtable}
\end{small}

As mentioned in Section \ref{dataset}, Bengali, Hindi, Sinhala, Nepali, Urdu, and Modern Standard Arabic are present twice, once in their respective native scripts and once in the Latin script. Chinese is also present twice, in Simplified and Traditional characters. There are 50 Indo-European languages, which we decide to display in smaller language families. Even so, Indo-Aryan is the most common language family (17), followed by Atlantic-Congo (16) and Afro-Asiatic (16). 

Note that the language code used is from \flores-200, and is not exactly the same as the older \flores-101 code, see the \href{https://github.com/facebookresearch/flores/blob/main/flores200/README.md}{\flores website} for details.

\subsection{Training Set} \label{trainset}
To create a training and development set to enable model finetuning to the task for adequate evaluation, we considered a diverse set of multiple-choice question-answering datasets. As there were little options for similarly-formatted datasets outside of English, we only selected English datasets.

After considering 33 different MRC datasets, we determine the most compatible to be RACE \cite{lai-etal-2017-race}, \textsc{SciQ} \cite{welbl-etal-2017-crowdsourcing}, \textsc{MultiRC} \cite{khashabi-etal-2018-looking}, \textsc{MCTest} \cite{richardson-etal-2013-mctest}, \textsc{MCScript2.0} \cite{ostermann-etal-2019-mcscript2}, and \textsc{ReClor} \cite{Yu2020ReClorAR}. For each of the 6 datasets, we unpack and restructure the passages and questions from their respective formats. Some required more reformatting, such as MCScript2.0, which only has two multiple choice options. For this dataset, we sampled answers from different questions to complete 4 candidate answers. For each, we then filter out less suitable samples, such as questions with multiple correct answers (as in MultiRC), excessively long passages, or fill-in-the-blank questions. We then created subgroups of questions based of surface characteristics such as passage length, question length, and topic. We experiment with these different strata to train the best \textsc{RoBERTa}-base model \cite{liu2019roberta} evaluated on the English set from \belebele in order to empirically validate the ability to teach a model the correct task. We use these empirical evaluations to finalize the training set, along with a development set of associated samples from the respective validation sets of the 6 above datasets. No test sets were included.

In the end, the dataset comprises 67.5k training samples and 3.7k development samples, more than half of which are from RACE. We provide a script at \href{https://github.com/facebookresearch/belebele}{
our github repo} to reconstruct this dataset for anyone to perform task finetuning.

Note: Since the training set is a joint sample of other datasets, it is governed by a different license than \belebele. Most importantly, we do not claim any of that work or datasets to be our own. See Appendix \ref{license}.

\subsection{Licensing} \label{license}

The \belebele dataset is licensed under CC-BY-SA, as is the case for the underlying \flores-200. Please refer to 
\href{https://github.com/facebookresearch/belebele}{
our github repo} for more information.

The training set and assembly code is, however, licensed differently. The majority of the training set (data and code) is licensed under CC-BY-NC, therefore the use of \belebele for commercial purposes requires a different data solution for model finetuning.

\subsection{Experiment Details} \label{evaldetails}
\subsubsection{Model fine-tuning}

As discussed in Section \ref{evalsettings}, for the fine-tuning of MLMs, we use the training set detailed in Appendix \ref{trainset}. For all settings, the training was performed using the HuggingFace transformers library. We use the development set for hyperparameter search and evaluate the two best training runs on the \belebele splits.

For fine-tuning on the English training set, we find that 3 or 4 epochs is optimal for performance on the development set.

For Translate-Train-All, we use machine translation on passages, questions, and answers \emph{separately}. Since there is almost 100x available data for this setting, we limit the training and validation sample to 650k and only train one epoch. We find that beyond this, there is not much improvement on the translated development set.

We provide fine-tuning hyperparameters of the reported runs below in Appendix \ref{section/xl_mlm_results}

\subsubsection{In-Context Learning Prompt}

As stated in Section \ref{evalsettings}, for 5-shot in-context learning, examples are sampled from the English training set and prompted to the model. The template used is as follows:

\texttt{P: <passage> \textbackslash{n} Q: <question> \textbackslash{n} A: <mc answer 1> \textbackslash{n} B: <mc answer 2> \textbackslash{n}  C: <mc answer 3> \textbackslash{n}  D: <mc answer 4> \textbackslash{n}  Answer: <Correct answer letter>}

Within the answers \{A, B, C, D\}, we determine the prediction to be the one with the highest probability (relative to the others). For all our results, we report the average score over 3 runs.

\subsubsection{Zero-Shot Instructions}
As stated in Section \ref{evalsettings}, we evaluate both GPT3.5 and \llamatwochat in the zero-shot setting by describing the task in natural language. We present the passage, question, and four possible answers, and instruct the model to provide the letter ``A'', ``B'', ``C'' or ``D '' as the answer. The instructions are given in English for all languages.  We perform post-processing steps and accept answers predicted as e.g. ``(A)'' instead of ``A''. The instructions and post-processing code are provided at \href{https://github.com/facebookresearch/belebele}{
our github repo}.

For the \emph{In-language Translated Instructions} setting, we replicate the above, except present \llamatwochat (70B) with instructions that are machine translated to the target language. We do not translate (or transliterate) the lettered answers we ask for (``A'', ``B'', ``C'' or ``D ''). Therefore, the post-processing steps are the same. 

For the \emph{Translate-Test} setting, we present the natural language instructions in English. The passage, questions, and answers, however, have been machine-translated \emph{individually} back to English.

\subsection{Annotation Guidelines}

\subsubsection{MCQA Annotation Guidelines} \label{guidelines}
The following is an abridged version of the particularized instructions provided to annotators for the task of creating a multiple-choice question-answering dataset. As mentioned, we additionally provided a positive and negative example for each guidelines to the annotators.

\begin{enumerate}
    \item Ensure that all answers, if not most, are decently plausible to require the test-taker to fully read and understand the passage. 
    \item In order to make the questions not overly easy, ensure that if the correct answer is word-for-word from the passage, at least some of the wrong answers are as well. This is to ensure that the person answering can’t just guess the right answer based off identifying the answer in the passage through skimming. 
    \item Make the questions as specific as possible, leave nothing to ambiguity.
    \item It should not be possible to answer the question without having read the passage, but the question must be answerable using just the passage (no external knowledge). We encourage the use of phrases such as “According to the passage…” in the questions if there may be ambiguity.
    \item Try to sometimes have the answers be word-for-word in the passage and for other questions, the answers be in your own words. We would like a balance of these two categories of question.
    \item Don’t write questions with double negatives that may trivially fool the person answering. 
    \item Try to ask questions in your own words and don’t copy and paste entire phrases from the paragraph. This allows us to truly evaluate the comprehension as opposed to recognizing patterns in the way the question is extracted from the text. That being said, making the questions comprehensible and including details is very important, as mentioned above.
    \item Avoid using ambiguous qualifiers (for example, “very” or “really”) or having the question centered around an extreme (for example, “the greatest” or “the most”). These phrases may leave room for ambiguity and subjectivity. In addition, qualified statements change with time and may not be valid in the future.
\end{enumerate}

\subsubsection{Translation Specifications} \label{specifications}
To align the dialect or langauge variant to \flores, we instructed the LSP to use the same localization as in the \flores-200 creation a few years prior. To align style, formality, and wording, we supplemented the traditional translation guidelines with the following:

\begin{displayquote}
Given that the associated translated passage is already established (and not subject to change as it is in a published dataset), the translations of the questions and answers have to be fully compatible with it. This means that for dates, proper nouns, units of measure, etc. where there is potential ambiguity, the translators have to follow what was done for the passage, even if they disagree that it is the more correct translation.

For example,

Hungarian Translated passage: “Jesus készen áll a vitára…”

English question: “What was Jesus working on when…”

Therefore, in the questions and answers Jesus must be translated as “Jesus” and not “Jézus”.
\end{displayquote}

\subsection{MCQA Lexical Featurization}
\label{lexical_features}

As discussed in Section~\ref{dataset}, we use lexical featurization to programmatically assess the ease of solvability of our multiple-choice questions. These features allow us to then run statistical tests and determine if there are implicit patterns (or "bias") that models can learn and exploit.

In the list below, note that lexical overlap was measured using 1-, 2-, 3-, and 4-grams.
\begin{enumerate}
    \item The frequency of lexical overlap between the passage and the correct answer, in comparison to overlap with wrong answers. This is the ensure the question cannot be answered by selecting the only options present in the original passage.
    \item The frequency of lexical overlap between the question and the correct answer, in comparison to overlap with wrong answers. This is the ensure the question cannot be answered by selecting options nearly extract-able from the question itself.
    \item If the question has lexical overlap with a sentence in the passage, is the correct answer in that same sentence, in comparison to overlap with wrong answers? This is the ensure the question cannot be answered by finding the sentence which the question is alluding to and extracting the answer from there directly.
    \item The frequency of a correct answer to have lexical overlap with the question, in comparison to wrong answers. Overlap with the question may giveaway the answer.
    \item The frequency of a correct answer to have lexical overlap with other answer options, in comparison to a pair of wrong answers. If the correct answer is more likely to have closer neighbors, it could allow models to narrow down options quicker.
    \item Simple length statistics (character and worth count) of correct answer, in comparison to wrong answers
\end{enumerate}

\subsection{Detailed Results Tables} \label{details}
\subsubsection{Cross-Lingual MLMs} \label{section/xl_mlm_results}
\begin{small}
    \begin{longtable}{ |c||c c c|c c c|} 
        \toprule
        \multicolumn{7}{c}{\textbf{Full Results for Cross-Lingual MLMs}} \\
        \midrule
        \textbf{Evaluation}  & \multicolumn{3}{|c|}{\textbf{Finetune in English}} & \multicolumn{3}{|c|}{\textbf{Translate-Train-All}} \\
        \textbf{Model Name} & \textbf{XLM-V} & \textbf{\textsc{InfoXLM}} & \textbf{XLM-R} & \textbf{XLM-V} & \textbf{\textsc{InfoXLM}} & \textbf{XLM-R}\\
        \textbf{Size/Variant} & \textbf{large} & \textbf{large} & \textbf{large} & \textbf{large} & \textbf{large} & \textbf{large} \\
        \midrule
        \textbf{AVG} & 55.6 & \textbf{56.2} & 54.0 & \textbf{60.2} & 60.0 & 58.9 \\ 
        \textbf{PCT Above 50} & \textbf{69.7\%} & 67.2\% & 64.8\% & \textbf{76.2\%} & 70.5\% & 69.7\% \\
        \textbf{PCT Above 70} & 21.9\% & \textbf{28.9\%} & 15.7\% & 33.1\% & \textbf{37.2\%} & 36.4\% \\ 
        \midrule
        eng\_Latn & 76.2 & \textbf{79.3} & 76.2 & 77.8 & \textbf{81.2} & 78.7 \\ 
        acm\_Arab & 51.2 & 57.3 & 55.4 & 55.3 & 57.6 & 59.2 \\ 
        afr\_Latn & 69.3 & 72.7 & 69.1 & 72.3 & 75.1 & 74.3 \\ 
        als\_Latn & 68.4 & 68.9 & 64.9 & 70.8 & 72.2 & 71.4 \\ 
        amh\_Ethi & 53.1 & 52.9 & 52.6 & 61.6 & 60.0 & 60.7 \\ 
        apc\_Arab & 56.1 & 58.8 & 57.9 & 57.7 & 60.6 & 61.9 \\ 
        arb\_Arab & 67.2 & 71.0 & 69.8 & 70.6 & 75.0 & 74.3 \\ 
        arb\_Latn & 29.3 & 32.2 & 27.6 & 31.6 & 33.4 & 30.6 \\ 
        ars\_Arab & 55.6 & 59.9 & 58.9 & 61.1 & 65.8 & 65.9 \\ 
        ary\_Arab & 43.8 & 48.7 & 44.0 & 48.0 & 52.8 & 52.6 \\ 
        arz\_Arab & 56.9 & 60.2 & 57.6 & 61.4 & 64.9 & 66.1 \\ 
        asm\_Beng & 53.7 & 53.6 & 49.3 & 58.6 & 58.8 & 56.9 \\ 
        azj\_Latn & 59.7 & 61.3 & 59.0 & 65.0 & 65.6 & 65.1 \\ 
        bam\_Latn & 34.2 & 34.9 & 33.2 & 39.2 & 39.1 & 36.9 \\ 
        ben\_Beng & 60.0 & 63.4 & 59.6 & 65.6 & 69.6 & 63.7 \\ 
        ben\_Latn & 46.8 & 36.9 & 38.8 & 53.0 & 42.7 & 48.1 \\ 
        bod\_Tibt & 24.0 & 24.9 & 23.7 & 24.8 & 23.3 & 36.9 \\ 
        bul\_Cyrl & 72.6 & 72.0 & 70.1 & 74.0 & 75.3 & 74.2 \\ 
        cat\_Latn & 71.6 & 74.4 & 72.0 & 75.7 & 78.1 & 74.7 \\ 
        ceb\_Latn & 45.4 & 44.1 & 42.3 & 52.0 & 52.6 & 50.7 \\ 
        ces\_Latn & 69.9 & 72.3 & 69.9 & 72.3 & 76.2 & 74.4 \\ 
        ckb\_Arab & 29.7 & 52.3 & 30.3 & 36.9 & 58.0 & 36.9 \\ 
        dan\_Latn & 70.8 & 74.1 & 72.9 & 73.0 & 76.3 & 74.7 \\ 
        deu\_Latn & 72.6 & 75.7 & 72.9 & 74.1 & 78.7 & 76.7 \\ 
        ell\_Grek & 70.3 & 72.3 & 70.3 & 73.1 & 74.9 & 73.0 \\ 
        est\_Latn & 63.2 & 67.2 & 64.8 & 68.7 & 70.7 & 70.4 \\ 
        eus\_Latn & 63.6 & 66.1 & 64.8 & 68.2 & 70.8 & 70.3 \\ 
        fin\_Latn & 69.1 & 72.4 & 72.2 & 73.0 & 75.2 & 74.9 \\ 
        fra\_Latn & 73.1 & 74.2 & 72.1 & 74.6 & 76.8 & 75.6 \\ 
        fuv\_Latn & 29.7 & 27.7 & 26.4 & 32.8 & 30.7 & 31.1 \\ 
        gaz\_Latn & 48.8 & 33.8 & 36.4 & 52.6 & 36.0 & 43.3 \\ 
        grn\_Latn & 53.9 & 37.8 & 37.9 & 59.6 & 40.6 & 41.9 \\ 
        guj\_Gujr & 58.7 & 57.0 & 54.1 & 63.3 & 65.9 & 63.1 \\ 
        hat\_Latn & 57.1 & 39.6 & 35.2 & 63.2 & 44.1 & 39.8 \\ 
        hau\_Latn & 51.0 & 41.1 & 48.2 & 53.4 & 48.1 & 53.0 \\ 
        heb\_Hebr & 67.2 & 68.2 & 64.8 & 69.3 & 72.3 & 70.6 \\ 
        hin\_Deva & 57.9 & 60.2 & 57.4 & 63.8 & 64.3 & 63.4 \\ 
        hin\_Latn & 53.1 & 49.7 & 46.8 & 57.6 & 55.4 & 58.9 \\ 
        hrv\_Latn & 70.0 & 72.4 & 69.9 & 71.2 & 75.3 & 74.0 \\ 
        hun\_Latn & 69.7 & 70.8 & 70.0 & 73.1 & 74.2 & 72.8 \\ 
        hye\_Armn & 59.4 & 61.0 & 58.9 & 65.9 & 66.1 & 64.7 \\ 
        ibo\_Latn & 40.1 & 32.2 & 31.2 & 46.8 & 32.0 & 32.2 \\ 
        ilo\_Latn & 37.4 & 36.3 & 33.8 & 38.1 & 40.6 & 39.7 \\ 
        ind\_Latn & 68.9 & 70.7 & 68.0 & 71.3 & 73.1 & 70.4 \\ 
        isl\_Latn & 67.3 & 66.0 & 63.8 & 70.1 & 68.9 & 69.0 \\ 
        ita\_Latn & 70.6 & 72.8 & 70.0 & 71.8 & 76.4 & 73.3 \\ 
        jav\_Latn & 64.2 & 59.8 & 60.8 & 67.2 & 63.3 & 66.8 \\ 
        jpn\_Jpan & 66.4 & 70.1 & 67.6 & 71.3 & 71.8 & 71.0 \\ 
        kac\_Latn & 32.0 & 29.1 & 32.1 & 33.8 & 34.0 & 33.3 \\ 
        kan\_Knda & 61.1 & 62.0 & 59.7 & 66.6 & 68.4 & 69.1 \\ 
        kat\_Geor & 64.7 & 64.8 & 63.6 & 68.0 & 68.9 & 67.4 \\ 
        kaz\_Cyrl & 60.1 & 61.6 & 56.8 & 64.9 & 65.3 & 64.7 \\ 
        kea\_Latn & 44.0 & 45.2 & 44.9 & 48.7 & 47.7 & 48.1 \\ 
        khk\_Cyrl & 56.7 & 58.8 & 57.8 & 61.1 & 64.6 & 64.2 \\ 
        khm\_Khmr & 60.0 & 59.0 & 57.7 & 63.0 & 64.2 & 63.8 \\ 
        kin\_Latn & 35.9 & 33.6 & 34.3 & 39.1 & 39.1 & 38.6 \\ 
        kir\_Cyrl & 65.4 & 63.4 & 61.8 & 68.3 & 68.2 & 67.7 \\ 
        kor\_Hang & 70.1 & 71.4 & 68.7 & 72.9 & 74.6 & 74.8 \\ 
        lao\_Laoo & 55.8 & 57.6 & 53.0 & 63.2 & 63.6 & 63.0 \\ 
        lin\_Latn & 44.7 & 33.2 & 30.6 & 50.9 & 35.3 & 34.4 \\ 
        lit\_Latn & 68.3 & 69.4 & 67.2 & 71.7 & 72.9 & 72.0 \\ 
        lug\_Latn & 39.9 & 29.4 & 31.6 & 47.8 & 34.7 & 34.7 \\ 
        luo\_Latn & 30.3 & 30.9 & 30.8 & 33.7 & 34.9 & 33.2 \\ 
        lvs\_Latn & 70.1 & 71.3 & 68.7 & 74.1 & 75.6 & 73.0 \\ 
        mal\_Mlym & 62.0 & 65.0 & 62.7 & 69.1 & 68.3 & 67.1 \\ 
        mar\_Deva & 62.6 & 65.2 & 60.8 & 69.2 & 68.8 & 67.2 \\ 
        mkd\_Cyrl & 67.8 & 69.3 & 65.7 & 71.0 & 73.8 & 72.8 \\ 
        mlt\_Latn & 37.9 & 57.1 & 38.1 & 40.2 & 63.7 & 42.7 \\ 
        mri\_Latn & 32.0 & 30.6 & 32.2 & 33.0 & 35.7 & 34.0 \\ 
        mya\_Mymr & 56.6 & 59.1 & 53.6 & 62.2 & 65.1 & 62.9 \\ 
        nld\_Latn & 68.4 & 71.7 & 71.0 & 68.6 & 74.0 & 72.8 \\ 
        nob\_Latn & 71.8 & 73.6 & 70.7 & 72.8 & 75.4 & 74.2 \\ 
        npi\_Deva & 58.4 & 60.7 & 55.7 & 64.4 & 65.8 & 62.7 \\ 
        npi\_Latn & 38.3 & 35.8 & 33.8 & 37.4 & 36.4 & 34.8 \\ 
        nso\_Latn & 45.9 & 31.3 & 30.0 & 53.2 & 34.1 & 34.7 \\ 
        nya\_Latn & 31.0 & 29.2 & 29.8 & 34.2 & 33.0 & 30.8 \\ 
        ory\_Orya & 60.8 & 62.1 & 58.6 & 65.6 & 65.4 & 63.9 \\ 
        pan\_Guru & 58.1 & 59.2 & 57.8 & 63.1 & 62.6 & 62.0 \\ 
        pbt\_Arab & 55.4 & 56.0 & 51.0 & 60.6 & 62.6 & 61.1 \\ 
        pes\_Arab & 68.3 & 69.1 & 68.2 & 70.8 & 73.6 & 72.0 \\ 
        plt\_Latn & 55.7 & 45.6 & 52.7 & 61.7 & 53.4 & 58.1 \\ 
        pol\_Latn & 69.0 & 70.4 & 67.4 & 72.1 & 73.7 & 72.7 \\ 
        por\_Latn & 70.9 & 74.3 & 70.6 & 73.8 & 77.1 & 74.0 \\ 
        ron\_Latn & 72.3 & 72.9 & 71.3 & 74.0 & 76.2 & 74.8 \\ 
        rus\_Cyrl & 71.9 & 73.8 & 72.2 & 75.4 & 76.8 & 77.1 \\ 
        shn\_Mymr & 26.9 & 25.2 & 26.3 & 25.0 & 26.4 & 27.0 \\ 
        sin\_Latn & 24.9 & 34.2 & 30.7 & 41.7 & 38.3 & 37.3 \\ 
        sin\_Sinh & 64.4 & 67.2 & 62.7 & 69.8 & 70.2 & 68.6 \\ 
        slk\_Latn & 69.3 & 71.9 & 70.2 & 72.6 & 76.7 & 73.0 \\ 
        slv\_Latn & 69.7 & 72.2 & 68.6 & 71.8 & 75.4 & 73.9 \\ 
        sna\_Latn & 34.8 & 37.2 & 33.2 & 37.1 & 38.6 & 35.9 \\ 
        snd\_Arab & 55.2 & 56.6 & 51.9 & 60.0 & 61.3 & 61.3 \\ 
        som\_Latn & 46.0 & 39.1 & 42.6 & 50.7 & 46.3 & 50.7 \\ 
        sot\_Latn & 46.8 & 29.3 & 31.3 & 52.0 & 31.9 & 32.7 \\ 
        spa\_Latn & 71.0 & 73.3 & 71.4 & 72.7 & 75.3 & 76.4 \\ 
        srp\_Cyrl & 71.0 & 70.9 & 71.1 & 73.6 & 76.1 & 75.9 \\ 
        ssw\_Latn & 39.8 & 30.6 & 34.3 & 47.1 & 34.3 & 38.9 \\ 
        sun\_Latn & 60.9 & 50.7 & 55.3 & 64.2 & 55.8 & 59.4 \\ 
        swe\_Latn & 73.0 & 75.0 & 74.2 & 74.2 & 76.9 & 75.1 \\ 
        swh\_Latn & 64.9 & 65.3 & 62.8 & 69.3 & 69.2 & 68.7 \\ 
        tam\_Taml & 61.8 & 64.6 & 61.7 & 67.4 & 69.4 & 65.3 \\ 
        tel\_Telu & 55.6 & 57.8 & 53.6 & 62.1 & 63.2 & 61.1 \\ 
        tgk\_Cyrl & 38.2 & 58.6 & 33.8 & 39.2 & 64.3 & 39.6 \\ 
        tgl\_Latn & 69.2 & 67.4 & 64.7 & 72.0 & 70.4 & 70.0 \\ 
        tha\_Thai & 63.8 & 68.1 & 65.8 & 69.0 & 68.9 & 70.1 \\ 
        tir\_Ethi & 33.3 & 36.7 & 33.8 & 39.9 & 42.1 & 37.7 \\ 
        tsn\_Latn & 49.0 & 35.0 & 30.8 & 49.8 & 35.7 & 34.3 \\ 
        tso\_Latn & 37.9 & 36.3 & 34.2 & 41.7 & 39.7 & 37.1 \\ 
        tur\_Latn & 66.7 & 70.2 & 66.8 & 70.6 & 72.0 & 72.0 \\ 
        ukr\_Cyrl & 70.4 & 70.9 & 71.0 & 72.3 & 74.9 & 75.0 \\ 
        urd\_Arab & 61.6 & 63.8 & 59.3 & 65.6 & 68.6 & 66.3 \\ 
        urd\_Latn & 42.2 & 42.6 & 40.8 & 49.4 & 48.9 & 48.4 \\ 
        uzn\_Latn & 65.2 & 66.9 & 64.4 & 69.1 & 70.6 & 70.2 \\ 
        vie\_Latn & 69.6 & 71.1 & 69.4 & 73.7 & 72.9 & 71.4 \\ 
        war\_Latn & 46.4 & 44.7 & 43.7 & 47.6 & 49.3 & 46.6 \\ 
        wol\_Latn & 36.8 & 32.2 & 30.4 & 40.6 & 32.3 & 32.2 \\ 
        xho\_Latn & 48.7 & 36.1 & 39.0 & 54.4 & 40.2 & 45.4 \\ 
        yor\_Latn & 35.0 & 29.3 & 28.7 & 38.6 & 32.0 & 27.9 \\ 
        zho\_Hans & 69.8 & 74.6 & 71.0 & 73.7 & 76.2 & 74.8 \\ 
        zho\_Hant & 69.2 & 72.4 & 67.1 & 73.1 & 74.3 & 71.3 \\ 
        zsm\_Latn & 69.1 & 72.6 & 69.9 & 72.4 & 73.3 & 72.2 \\ 
        zul\_Latn & 46.9 & 36.4 & 39.0 & 54.2 & 39.8 & 44.1 \\
        \bottomrule
\caption{Results of Cross-Lingual MLMs in the two settings described in Section \ref{experiments}.}
\label{table:xlm_results}
\end{longtable}
\end{small}

For all, the \emph{large} version was used which is the same architecture across all three. XLM-V has a significantly larger vocabulary size, leading to more total parameters. We find that in general, \textsc{InfoXLM} and XLM-V are very similar and both out-perform XLM-R across the board. \textsc{InfoXLM} outperforms XLM-V in higher- and medium-resource languages, while XLM-V performs better on the lowest-resource languages. As a result, XLM-V has the most scores above $50$, but \textsc{InfoXLM} has more scores above $70$. In the below table, we provide the hyperparameters used for these specific runs.

\begin{small}
    \begin{longtable}{ |c||c c c|c c c|} 
        \toprule
        \multicolumn{7}{c}{\textbf{Full Results for Cross-Lingual MLMs}} \\
        \midrule
        \textbf{Evaluation}  & \multicolumn{3}{|c|}{\textbf{Finetune in English}} & \multicolumn{3}{|c|}{\textbf{Translate-Train-All}} \\
        \textbf{Model Name} & \textbf{XLM-V} & \textbf{\textsc{InfoXLM}} & \textbf{XLM-R} & \textbf{XLM-V} & \textbf{\textsc{InfoXLM}} & \textbf{XLM-R}\\
        \textbf{Size/Variant} & \textbf{large} & \textbf{large} & \textbf{large} & \textbf{large} & \textbf{large} & \textbf{large} \\
        \midrule
        epochs & 3 & 4 & 3 & 1 & 1 & 1 \\
        training set size & 67.5k & 67.5k & 67.5k & 650k & 650k & 650k \\
        learning rate & 5e-6 & 4e-6 & 5e-6 & 3-6 & 3e-6 & 3e-6 \\
        weight decay & 0.01 & 0.01 & 0.01 & 0.001 & 0.001 & 0.001 \\
        batch size & 64 & 64 & 64 & 64 & 64 & 64 \\
        \bottomrule
\caption{Finetuning Hyperparameters for the runs reported above. More details in Appendix \ref{evaldetails}.}
\label{finetuning_hyperparams}
\end{longtable}
\end{small}

\subsubsection{LLMs}
\begin{small}
    \begin{longtable}{ |c||c c|c c c|c|} 
    \toprule
        \multicolumn{7}{c}{\textbf{Full Results for Large Language Models}} \\
        \midrule
        \textbf{Evaluation}  & \multicolumn{2}{|c|}{\textbf{Zero-Shot for Instructed Models}} & \multicolumn{3}{|c|}{\textbf{5-shot In-Context Learning}} & \textbf{Translate-Train-All}\\
        \textbf{Model Name} & \textbf{\gptturbo}  & \textbf{\llamatwochat} & \textbf{\llama 2} & \textbf{\llama 1}  & \textbf{\falcon} & \textbf{XLM-V}\\
        \textbf{Size/Variant} & & \textbf{70B} & \textbf{70B} & \textbf{65B} & \textbf{40B} & \textbf{large} \\
        \midrule
        \textbf{AVG} & \textbf{50.6} & 41.5 & \textbf{48.0} & 40.9 & 37.3 & 60.2 \\ 
        \textbf{PCT Above 50} & \textbf{43.4}\% & 27.1\% & \textbf{38.5}\% & 25.4\% & 16.4\% & 76.2\% \\ 
        \textbf{PCT Above 70} & \textbf{28.9}\% & 2.5\% & \textbf{26.2}\% & 12.3\% & 1.6\% & 33.1\% \\
        \midrule
        eng\_Latn & \textbf{87.7} & 78.8 & \textbf{90.9} & 82.5 & 77.2 & 77.8 \\ 
        acm\_Arab & 51.6 & 35.9 & 47.9 & 37.9 & 37.6 & 55.3 \\ 
        afr\_Latn & 78.3 & 57.9 & 75.9 & 60.7 & 53.4 & 72.3 \\ 
        als\_Latn & 67.1 & 36.0 & 45.4 & 34.9 & 36.6 & 70.8 \\ 
        amh\_Ethi & 28.7 & 28.9 & 27.5 & 27.8 & 24.8 & 61.6 \\ 
        apc\_Arab & 55.6 & 38.8 & 51.2 & 39.6 & 36.3 & 57.7 \\ 
        arb\_Arab & 69.3 & 42.3 & 61.7 & 44.1 & 38.3 & 70.6 \\ 
        arb\_Latn & 31.1 & 30.2 & 26.8 & 28.0 & 26.3 & 31.6 \\ 
        ars\_Arab & 55.1 & 37.4 & 50.2 & 40.7 & 32.1 & 61.1 \\ 
        ary\_Arab & 45.7 & 32.6 & 40.6 & 33.1 & 32.3 & 48.0 \\ 
        arz\_Arab & 56.7 & 37.3 & 50.7 & 37.4 & 33.0 & 61.4 \\ 
        asm\_Beng & 36.0 & 35.7 & 32.3 & 28.9 & 22.4 & 58.6 \\ 
        azj\_Latn & 54.9 & 33.4 & 42.2 & 33.6 & 34.1 & 65.0 \\ 
        bam\_Latn & 31.7 & 29.4 & 30.3 & 28.4 & 29.7 & 39.2 \\ 
        ben\_Beng & 43.6 & 34.9 & 39.1 & 33.4 & 22.6 & 65.6 \\ 
        ben\_Latn & 34.6 & 30.4 & 29.6 & 29.2 & 32.1 & 53.0 \\ 
        bod\_Tibt & 26.6 & 28.3 & 25.7 & 24.9 & 26.8 & 24.8 \\ 
        bul\_Cyrl & 76.0 & 65.0 & 80.4 & 69.3 & 41.9 & 74.0 \\ 
        cat\_Latn & 78.4 & 68.2 & 84.6 & 76.3 & 58.8 & 75.7 \\ 
        ceb\_Latn & 53.3 & 40.6 & 50.4 & 38.9 & 39.2 & 52.0 \\ 
        ces\_Latn & 76.9 & 65.0 & 81.1 & 70.7 & 65.0 & 72.3 \\ 
        ckb\_Arab & 31.8 & 32.8 & 28.7 & 31.6 & 28.9 & 36.9 \\ 
        dan\_Latn & 80.7 & 66.2 & 83.6 & 73.6 & 56.2 & 73.0 \\ 
        deu\_Latn & 83.3 & 69.4 & 84.6 & 76.0 & 70.1 & 74.1 \\ 
        ell\_Grek & 73.0 & 50.7 & 64.9 & 44.2 & 31.2 & 73.1 \\ 
        est\_Latn & 73.1 & 36.6 & 53.0 & 36.3 & 34.9 & 68.7 \\ 
        eus\_Latn & 40.9 & 31.1 & 34.7 & 32.8 & 38.9 & 68.2 \\ 
        fin\_Latn & 77.9 & 62.7 & 79.3 & 55.7 & 42.8 & 73.0 \\ 
        fra\_Latn & 83.1 & 72.2 & 86.4 & 77.5 & 69.7 & 74.6 \\ 
        fuv\_Latn & 26.1 & 29.8 & 24.9 & 25.4 & 25.1 & 32.8 \\ 
        gaz\_Latn & 30.3 & 29.3 & 27.8 & 29.1 & 24.9 & 52.6 \\ 
        grn\_Latn & 34.2 & 32.2 & 32.4 & 30.3 & 33.8 & 59.6 \\ 
        guj\_Gujr & 38.4 & 31.1 & 27.1 & 25.7 & 24.7 & 63.3 \\ 
        hat\_Latn & 51.6 & 34.1 & 37.4 & 33.7 & 36.2 & 63.2 \\ 
        hau\_Latn & 32.2 & 32.1 & 28.0 & 26.4 & 28.9 & 53.4 \\ 
        heb\_Hebr & 64.2 & 41.4 & 54.9 & 41.4 & 31.1 & 69.3 \\ 
        hin\_Deva & 49.1 & 42.0 & 52.6 & 38.4 & 27.1 & 63.8 \\ 
        hin\_Latn & 52.3 & 39.2 & 49.0 & 34.2 & 40.0 & 57.6 \\ 
        hrv\_Latn & 78.4 & 64.7 & 79.8 & 66.9 & 48.7 & 71.2 \\ 
        hun\_Latn & 74.6 & 61.1 & 78.8 & 66.7 & 37.7 & 73.1 \\ 
        hye\_Armn & 35.0 & 31.9 & 34.1 & 32.1 & 25.4 & 65.9 \\ 
        ibo\_Latn & 28.4 & 30.1 & 27.4 & 25.3 & 30.2 & 46.8 \\ 
        ilo\_Latn & 37.1 & 33.2 & 36.6 & 32.1 & 35.1 & 38.1 \\ 
        ind\_Latn & 74.2 & 61.3 & 81.4 & 55.7 & 52.1 & 71.3 \\ 
        isl\_Latn & 62.3 & 38.0 & 54.3 & 42.1 & 36.4 & 70.1 \\ 
        ita\_Latn & 80.0 & 68.6 & 84.5 & 76.1 & 66.4 & 71.8 \\ 
        jav\_Latn & 46.7 & 37.0 & 40.3 & 33.0 & 36.8 & 67.2 \\ 
        jpn\_Jpan & 70.9 & 56.6 & 77.6 & 53.9 & 49.6 & 71.3 \\ 
        kac\_Latn & 30.9 & 30.7 & 27.7 & 28.6 & 27.8 & 33.8 \\ 
        kan\_Knda & 40.6 & 21.9 & 25.7 & 24.4 & 24.0 & 66.6 \\ 
        kat\_Geor & 33.0 & 34.6 & 37.8 & 34.3 & 23.4 & 68.0 \\ 
        kaz\_Cyrl & 35.0 & 32.4 & 29.3 & 32.4 & 32.6 & 64.9 \\ 
        kea\_Latn & 46.0 & 38.1 & 45.4 & 38.1 & 38.0 & 48.7 \\ 
        khk\_Cyrl & 32.0 & 31.1 & 29.8 & 28.4 & 27.4 & 61.1 \\ 
        khm\_Khmr & 30.4 & 30.6 & 27.0 & 28.2 & 25.0 & 63.0 \\ 
        kin\_Latn & 35.2 & 30.6 & 29.8 & 28.5 & 31.9 & 39.1 \\ 
        kir\_Cyrl & 37.9 & 32.2 & 34.6 & 32.5 & 31.9 & 68.3 \\ 
        kor\_Hang & 67.1 & 56.3 & 77.8 & 52.9 & 40.2 & 72.9 \\ 
        lao\_Laoo & 30.0 & 26.5 & 24.3 & 26.2 & 28.1 & 63.2 \\ 
        lin\_Latn & 33.8 & 31.0 & 28.0 & 30.4 & 29.3 & 50.9 \\ 
        lit\_Latn & 72.0 & 39.7 & 52.1 & 39.6 & 39.3 & 71.7 \\ 
        lug\_Latn & 28.4 & 30.9 & 29.2 & 28.3 & 28.9 & 47.8 \\ 
        luo\_Latn & 27.1 & 31.2 & 29.4 & 29.3 & 29.9 & 33.7 \\ 
        lvs\_Latn & 70.8 & 41.0 & 51.3 & 39.0 & 37.6 & 74.1 \\ 
        mal\_Mlym & 34.9 & 30.1 & 32.4 & 30.0 & 21.2 & 69.1 \\ 
        mar\_Deva & 38.3 & 34.8 & 41.2 & 32.9 & 25.0 & 69.2 \\ 
        mkd\_Cyrl & 69.4 & 55.7 & 72.5 & 56.2 & 38.1 & 71.0 \\ 
        mlt\_Latn & 44.8 & 36.2 & 44.9 & 36.7 & 35.4 & 40.2 \\ 
        mri\_Latn & 33.3 & 31.8 & 28.5 & 32.0 & 29.7 & 33.0 \\ 
        mya\_Mymr & 30.3 & 31.3 & 24.1 & 24.2 & 22.6 & 62.2 \\ 
        nld\_Latn & 80.4 & 66.2 & 82.2 & 73.3 & 66.7 & 68.6 \\ 
        nob\_Latn & 79.0 & 65.7 & 81.8 & 70.9 & 60.8 & 72.8 \\ 
        npi\_Deva & 40.4 & 32.9 & 40.4 & 33.0 & 25.4 & 64.4 \\ 
        npi\_Latn & 35.1 & 30.4 & 30.2 & 30.0 & 30.9 & 37.4 \\ 
        nso\_Latn & 33.6 & 30.1 & 30.4 & 27.4 & 29.3 & 53.2 \\ 
        nya\_Latn & 33.2 & 29.3 & 27.3 & 28.7 & 29.3 & 34.2 \\ 
        ory\_Orya & ~ & 29.2 & 24.8 & 23.9 & 23.7 & 65.6 \\ 
        pan\_Guru & 39.1 & 33.1 & 26.3 & 27.1 & 23.4 & 63.1 \\ 
        pbt\_Arab & 32.3 & 30.2 & 30.8 & 29.4 & 29.4 & 60.6 \\ 
        pes\_Arab & 61.8 & 41.8 & 53.9 & 41.0 & 35.9 & 70.8 \\ 
        plt\_Latn & 32.3 & 30.5 & 29.6 & 31.0 & 31.4 & 61.7 \\ 
        pol\_Latn & 74.7 & 61.7 & 79.2 & 67.0 & 59.9 & 72.1 \\ 
        por\_Latn & 83.0 & 70.2 & 86.1 & 75.4 & 68.3 & 73.8 \\ 
        ron\_Latn & 77.4 & 65.6 & 83.4 & 73.2 & 66.6 & 74.0 \\ 
        rus\_Cyrl & 78.4 & 67.0 & 82.7 & 73.1 & 48.1 & 75.4 \\ 
        shn\_Mymr & ~ & 28.2 & 25.6 & 22.7 & 24.0 & 25.0 \\ 
        sin\_Latn & 30.4 & 31.9 & 33.8 & 27.9 & 32.6 & 41.7 \\ 
        sin\_Sinh & 32.6 & 33.4 & 25.2 & 29.4 & 27.7 & 69.8 \\ 
        slk\_Latn & 77.3 & 58.8 & 75.2 & 60.4 & 57.0 & 72.6 \\ 
        slv\_Latn & 77.4 & 62.4 & 76.7 & 65.6 & 43.7 & 71.8 \\ 
        sna\_Latn & 35.4 & 30.2 & 27.4 & 28.3 & 31.6 & 37.1 \\ 
        snd\_Arab & 34.1 & 29.7 & 30.9 & 28.9 & 30.2 & 60.0 \\ 
        som\_Latn & 32.4 & 30.3 & 27.8 & 27.6 & 29.9 & 50.7 \\ 
        sot\_Latn & 33.9 & 30.0 & 28.9 & 26.8 & 29.9 & 52.0 \\ 
        spa\_Latn & 79.2 & 68.4 & 85.0 & 74.8 & 69.2 & 72.7 \\ 
        srp\_Cyrl & 74.8 & 65.1 & 81.0 & 70.7 & 40.2 & 73.6 \\ 
        ssw\_Latn & 32.0 & 30.7 & 27.7 & 28.0 & 30.1 & 47.1 \\ 
        sun\_Latn & 38.9 & 34.9 & 37.8 & 30.7 & 34.1 & 64.2 \\ 
        swe\_Latn & 81.7 & 67.4 & 82.7 & 73.7 & 67.3 & 74.2 \\ 
        swh\_Latn & 70.3 & 35.1 & 39.6 & 34.4 & 36.7 & 69.3 \\ 
        tam\_Taml & 32.8 & 34.4 & 33.2 & 31.6 & 24.4 & 67.4 \\ 
        tel\_Telu & 34.6 & 27.5 & 25.9 & 26.6 & 22.4 & 62.1 \\ 
        tgk\_Cyrl & 37.7 & 32.5 & 34.0 & 33.1 & 32.7 & 39.2 \\ 
        tgl\_Latn & 66.7 & 49.6 & 68.1 & 48.3 & 47.7 & 72.0 \\ 
        tha\_Thai & 55.7 & 38.9 & 46.2 & 35.0 & 33.0 & 69.0 \\ 
        tir\_Ethi & 28.4 & 29.6 & 24.5 & 23.5 & 25.0 & 39.9 \\ 
        tsn\_Latn & 31.8 & 30.1 & 28.5 & 24.7 & 31.2 & 49.8 \\ 
        tso\_Latn & 33.4 & 30.0 & 30.4 & 28.0 & 29.7 & 41.7 \\ 
        tur\_Latn & 69.9 & 47.3 & 65.4 & 42.1 & 39.6 & 70.6 \\ 
        ukr\_Cyrl & 72.8 & 65.7 & 80.8 & 69.7 & 41.9 & 72.3 \\ 
        urd\_Arab & 48.3 & 37.0 & 43.2 & 34.7 & 31.7 & 65.6 \\ 
        urd\_Latn & 40.3 & 34.1 & 38.0 & 30.1 & 34.2 & 49.4 \\ 
        uzn\_Latn & 44.1 & 33.1 & 35.1 & 30.6 & 33.1 & 69.1 \\ 
        vie\_Latn & 72.9 & 59.6 & 78.4 & 43.5 & 41.4 & 73.7 \\ 
        war\_Latn & 48.9 & 39.3 & 44.4 & 37.4 & 38.6 & 47.6 \\ 
        wol\_Latn & 29.0 & 28.9 & 27.6 & 26.0 & 26.8 & 40.6 \\ 
        xho\_Latn & 30.0 & 29.9 & 28.2 & 27.6 & 30.2 & 54.4 \\ 
        yor\_Latn & 29.1 & 30.1 & 28.3 & 27.7 & 27.2 & 38.6 \\ 
        zho\_Hans & 77.6 & 62.4 & 83.7 & 64.6 & 66.0 & 73.7 \\ 
        zho\_Hant & 76.3 & 59.3 & 82.0 & 57.7 & 62.2 & 73.1 \\ 
        zsm\_Latn & 74.0 & 56.4 & 76.3 & 51.7 & 51.3 & 72.4 \\ 
        zul\_Latn & 30.4 & 30.2 & 29.7 & 27.1 & 30.7 & 54.2 \\  
    \bottomrule
  
    \caption{Results on LLMs, with comparison to full finetuning on XLM-V}
    \label{table:llm_results}
    \end{longtable}
\end{small}

The evaluation settings and models are described in more detail in Section \ref{experiments}. We see that none of these models can understand many of the 122 languages, while demonstrating excellent performance on high-resource languages. The 175B-parameter \gptturbo outperforms \llamatwochat (70B) across the board and has comparable results to \llama 2 (70B) even though it is in zero-shot. Note that \gptturbo threw errors when processing characters in Shan (shn\_Mymr) and Oriya (ory\_Orya) and therefore we could not evaluate the results. For the purposes of aggregated scores, we consider this a score of 25.0. For comparison to fully-finetuned multilingual models, we re-provide the results of XLM-V-large.

\subsubsection{Languages in Multiple Scripts} \label{scripts_appendix}
\begin{small}
    \begin{longtable}{ c|c c c c| c}
        \toprule
        \multicolumn{6}{c}{\textbf{Comparative Results for Languages with Multiple Scripts}} \\
        \midrule
        \textbf{Evaluation}  & \textbf{Zero-Shot} & \multicolumn{2}{c}{\textbf{Five-Shot}} & \textbf{Finetune in English} &  \\
        \textbf{Model Name} & \textbf{\gptturbo}  & \textbf{\llama 2 (70B)} & \textbf{\falcon (40B)} & \textbf{\textsc{InfoXLM}-large} & \textbf{AVG}\\
        \midrule
        arb\_Arab & \textbf{69.3} & \textbf{61.7} & \textbf{38.3} & \textbf{71.0} & \textbf{60.1} \\ 
        arb\_Latn & 31.1 & 26.8 & 26.3 & 32.2 & 29.1 \\ 
        \midrule
        ben\_Beng & \textbf{43.6} & \textbf{39.1} & 22.6 & \textbf{63.4} & \textbf{42.2} \\ 
        ben\_Latn & 34.6 & 29.6 & \textbf{32.1} & 36.9 & 33.3 \\ 
        \midrule
        hin\_Deva & 49.1 & \textbf{52.6} & 27.1 & \textbf{60.2} & 47.3 \\ 
        hin\_Latn & \textbf{52.3} & 49.0 & \textbf{40.0} & 49.7 & 47.8 \\ 
        \midrule
        npi\_Deva & \textbf{40.4} & \textbf{40.4} & 25.4 & \textbf{60.7} & \textbf{41.7} \\ 
        npi\_Latn & 35.1 & 30.2 & \textbf{30.9} & 35.8 & 33.0 \\ 
        \midrule
        sin\_Sinh & 32.6 & 25.2 & 27.7 & \textbf{67.2} & \textbf{38.2} \\ 
        sin\_Latn & 30.4 & \textbf{33.8} & \textbf{32.6} & 34.2 & 32.8 \\ 
        \midrule
        urd\_Arab & \textbf{48.3} & \textbf{43.2} & 31.7 & \textbf{63.8} & \textbf{46.7} \\ 
        urd\_Latn & 40.3 & 38.0 & \bf 34.2 & 42.6 & 38.8 \\ 
        \midrule
        zho\_Hant & 76.3 & 82.0 & 62.2 & 72.4 & 73.3 \\ 
        zho\_Hans & 77.6 & 83.7 & \textbf{66.0} & \textbf{74.6} & \textbf{75.5} \\ 
        \bottomrule
    \caption{Selected results from 3 models in differing settings comparing languages present in multiple scripts.}
    \label{table:scripts}
    \end{longtable}
\end{small}

We find that generally, the performance is higher in the native script than the romanized version, except for \falcon which displays the opposite trend on the 5 Indo-Aryan languages. In Chinese, performance on simplified \& traditional are very similar with simplified being higher across all 4. For \textsc{InfoXLM}, we display the English finetuning score.

\subsubsection{Translate-Test} \label{tt_appendix}
\begin{small}
    \begin{longtable}{ |c||c c | c|} 
    \toprule
        \multicolumn{4}{c}{\textbf{Translate-Test Results on 91 languages}} \\
        \midrule
        \textbf{Evaluation}  & \textbf{Zero-Shot} & \textbf{Translate-Test, Zero-Shot} & \textbf{Translate-Train-All} \\
        
        \textbf{Model} & \multicolumn{2}{|c|}{\textbf{\llamatwochat (70B)}} & \textbf{XLM-V-large} \\
        \midrule
        \textbf{AVG} & 44.0 & \textbf{57.1} & 64.9 \\ 
        \textbf{PCT Above 50} & 35.2\% & \textbf{78.0\%} & 90.1\% \\
        \textbf{PCT Above 70} & 2.2\% & 2.2\% & 42.9\% \\
        eng\_Latn & 78.8 & 78.8 & 77.8 \\
        \midrule
        fra\_Latn & 72.2 & 70.6 & 73.1 \\ 
        por\_Latn & 70.2 & 69.9 & 70.9 \\ 
        deu\_Latn & 69.4 & 65.7 & 72.6 \\ 
        ita\_Latn & 68.6 & 66.1 & 70.6 \\ 
        spa\_Latn & 68.4 & 69.3 & 71.0 \\ 
        cat\_Latn & 68.2 & 67.0 & 71.6 \\ 
        swe\_Latn & 67.4 & 66.1 & 73.0 \\ 
        rus\_Cyrl & 67.0 & 67.3 & 71.9 \\ 
        dan\_Latn & 66.2 & 66.8 & 70.8 \\ 
        nld\_Latn & 66.2 & 67.2 & 68.4 \\ 
        nob\_Latn & 65.7 & 68.3 & 71.8 \\ 
        ukr\_Cyrl & 65.7 & 66.0 & 70.4 \\ 
        ron\_Latn & 65.6 & 67.0 & 72.3 \\ 
        srp\_Cyrl & 65.1 & 66.2 & 71.0 \\ 
        bul\_Cyrl & 65.0 & 67.7 & 72.6 \\ 
        ces\_Latn & 65.0 & 65.6 & 69.9 \\ 
        hrv\_Latn & 64.7 & 65.3 & 70.0 \\ 
        fin\_Latn & 62.7 & 61.1 & 69.1 \\ 
        slv\_Latn & 62.4 & 61.2 & 69.7 \\ 
        zho\_Hans & 62.4 & 71.2 & 69.8 \\ 
        pol\_Latn & 61.7 & 63.0 & 69.0 \\ 
        ind\_Latn & 61.3 & 64.8 & 68.9 \\ 
        hun\_Latn & 61.1 & 62.9 & 69.7 \\ 
        vie\_Latn & 59.6 & 59.4 & 69.6 \\ 
        zho\_Hant & 59.3 & 65.8 & 69.2 \\ 
        slk\_Latn & 58.8 & 66.2 & 69.3 \\ 
        afr\_Latn & 57.9 & 65.0 & 69.3 \\ 
        jpn\_Jpan & 56.6 & 54.8 & 66.4 \\ 
        zsm\_Latn & 56.4 & 67.0 & 69.1 \\ 
        kor\_Hang & 56.3 & 56.7 & 70.1 \\ 
        mkd\_Cyrl & 55.7 & 66.7 & 67.8 \\ 
        ell\_Grek & 50.7 & 67.6 & 70.3 \\ 
        tgl\_Latn & 49.6 & 62.2 & 69.2 \\ 
        tur\_Latn & 47.3 & 62.6 & 66.7 \\ 
        arb\_Arab & 42.3 & 60.7 & 67.2 \\ 
        hin\_Deva & 42.0 & 62.8 & 57.9 \\ 
        pes\_Arab & 41.8 & 59.6 & 68.3 \\ 
        heb\_Hebr & 41.4 & 62.0 & 67.2 \\ 
        lvs\_Latn & 41.0 & 60.9 & 70.1 \\ 
        ceb\_Latn & 40.6 & 62.6 & 45.4 \\ 
        lit\_Latn & 39.7 & 60.8 & 68.3 \\ 
        hin\_Latn & 39.2 & 52.7 & 53.1 \\ 
        tha\_Thai & 38.9 & 54.1 & 63.8 \\ 
        isl\_Latn & 38.0 & 58.1 & 67.3 \\ 
        jav\_Latn & 37.0 & 55.3 & 64.2 \\ 
        urd\_Arab & 37.0 & 59.4 & 61.6 \\ 
        est\_Latn & 36.6 & 59.4 & 63.2 \\ 
        als\_Latn & 36.0 & 63.1 & 68.4 \\ 
        asm\_Beng & 35.7 & 57.7 & 53.7 \\ 
        swh\_Latn & 35.1 & 57.8 & 64.9 \\ 
        ben\_Beng & 34.9 & 61.0 & 60.0 \\ 
        sun\_Latn & 34.9 & 50.8 & 60.9 \\ 
        mar\_Deva & 34.8 & 60.0 & 62.6 \\ 
        kat\_Geor & 34.6 & 57.7 & 64.7 \\ 
        tam\_Taml & 34.4 & 55.9 & 61.8 \\ 
        urd\_Latn & 34.1 & 43.0 & 42.2 \\ 
        hat\_Latn & 34.1 & 56.3 & 57.1 \\ 
        azj\_Latn & 33.4 & 55.6 & 59.7 \\ 
        sin\_Sinh & 33.4 & 57.7 & 64.4 \\ 
        pan\_Guru & 33.1 & 57.6 & 58.1 \\ 
        npi\_Deva & 32.9 & 62.0 & 58.4 \\ 
        ckb\_Arab & 32.8 & 51.3 & 29.7 \\ 
        kaz\_Cyrl & 32.4 & 53.2 & 60.1 \\ 
        hau\_Latn & 32.1 & 43.4 & 51.0 \\ 
        hye\_Armn & 31.9 & 58.0 & 59.4 \\ 
        mya\_Mymr & 31.3 & 46.6 & 56.6 \\ 
        khk\_Cyrl & 31.1 & 52.2 & 56.7 \\ 
        guj\_Gujr & 31.1 & 59.6 & 58.7 \\ 
        lin\_Latn & 31.0 & 40.3 & 44.7 \\ 
        lug\_Latn & 30.9 & 38.7 & 39.9 \\ 
        ssw\_Latn & 30.7 & 43.2 & 39.8 \\ 
        khm\_Khmr & 30.6 & 52.8 & 60.0 \\ 
        plt\_Latn & 30.5 & 46.7 & 55.7 \\ 
        ben\_Latn & 30.4 & 45.1 & 46.8 \\ 
        som\_Latn & 30.3 & 40.8 & 46.0 \\ 
        pbt\_Arab & 30.2 & 48.8 & 55.4 \\ 
        zul\_Latn & 30.2 & 44.4 & 46.9 \\ 
        nso\_Latn & 30.1 & 43.4 & 45.9 \\ 
        tsn\_Latn & 30.1 & 40.4 & 49.0 \\ 
        yor\_Latn & 30.1 & 37.7 & 35.0 \\ 
        ibo\_Latn & 30.1 & 35.3 & 40.1 \\ 
        mal\_Mlym & 30.1 & 63.0 & 62.0 \\ 
        xho\_Latn & 29.9 & 49.2 & 48.7 \\ 
        fuv\_Latn & 29.8 & 29.4 & 29.7 \\ 
        gaz\_Latn & 29.3 & 37.0 & 48.8 \\ 
        ory\_Orya & 29.2 & 57.8 & 60.8 \\ 
        amh\_Ethi & 28.9 & 50.4 & 53.1 \\ 
        wol\_Latn & 28.9 & 39.0 & 36.8 \\ 
        tel\_Telu & 27.5 & 54.3 & 55.6 \\ 
        lao\_Laoo & 26.5 & 47.4 & 55.8 \\ 
        kan\_Knda & 21.9 & 62.0 & 61.1 \\ 
\bottomrule
    \caption{Comparing \llamatwochat Zero-Shot performance In-Language vs Translate-Test on 91 languages, with an aditional comparison to Translate-Train-All finetuning on XLM-V.}
    \label{table:translate_test}
    \end{longtable}
\end{small}

We evaluate 91 of the \nlangs languages in Translate-Test and find that Translate-Test performance dominates in-language performance on a big majority of languages for \llamatwochat (70B) in zero-shot. A few head languages such as German and Italian have higher scores in the traditional setting, but nearly all medium-resource languages are better understood with machine translation. For nearly all low-resource languages, the difference is over 20 accuracy points. For comparison, we see that machine translation lifts \llamatwochat performance in zero-shot not far from fully finetuned XLM-V-large. This is illustrated more clearly in (Fig. \ref{translate_test_fig}) below. 

\begin{figure}[H]
    \centering
    \includegraphics[width=15cm]{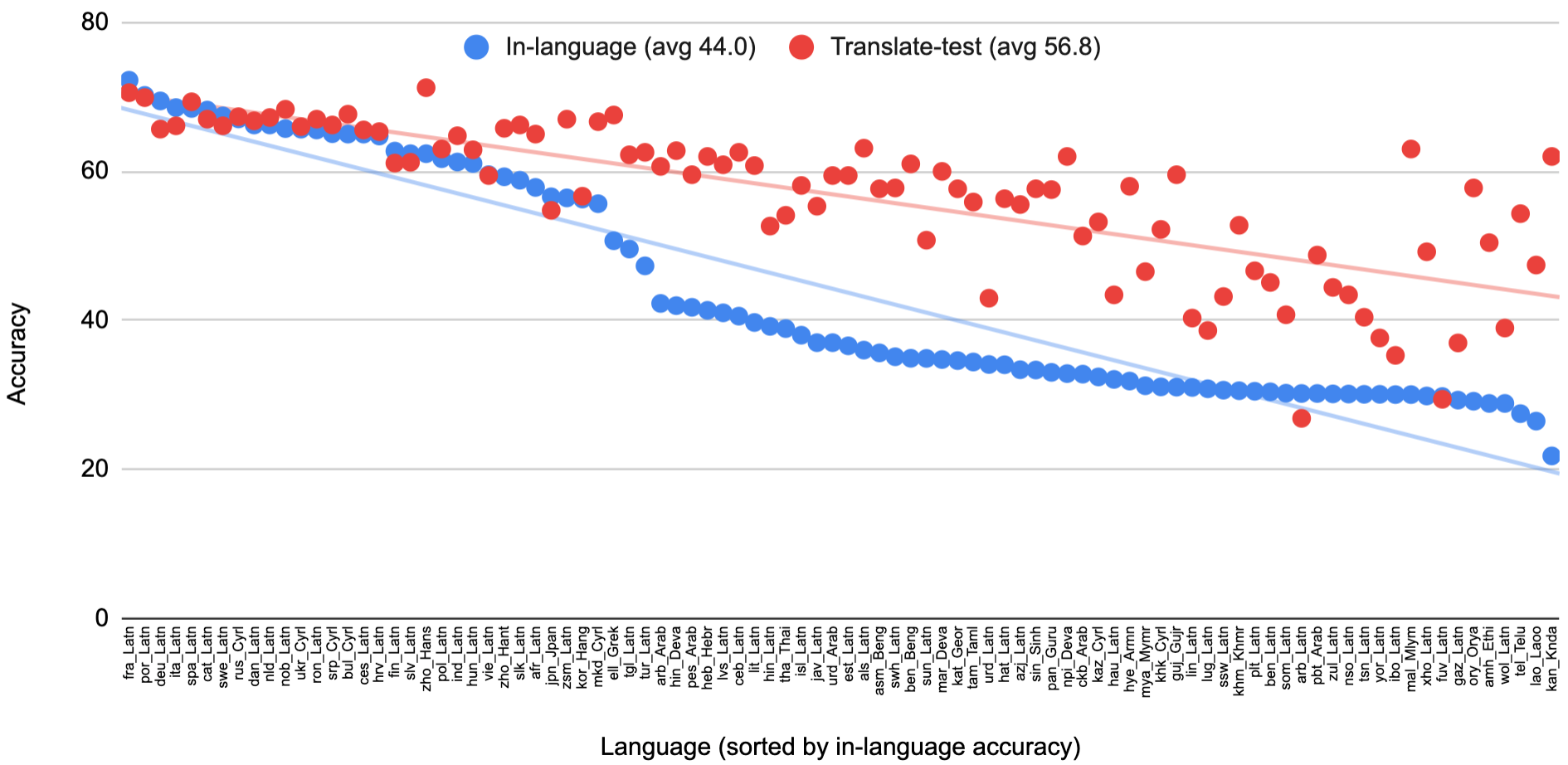}
    \caption{Comparison of \llamatwochat (70B) zero-shot performance on Translate-Test and the standard in-language evaluation.}
    \label{translate_test_fig}
\end{figure}

\end{document}